\documentclass[lettersize,journal]{IEEEtran}
\usepackage{amsmath,amsfonts}
\usepackage{algorithmic}
\usepackage{algorithm}
\usepackage{array}
\usepackage[caption=false,font=normalsize,labelfont=sf,textfont=sf]{subfig}
\usepackage{textcomp}
\usepackage{stfloats}
\usepackage{url}
\usepackage{verbatim}
\usepackage{graphicx}
\usepackage{cite}
\usepackage[numbers,sort&compress]{natbib}
\usepackage{color}

\usepackage{amssymb}
\makeatletter

\newcommand{\Rmnum}[1]{\expandafter\@slowromancap\romannumeral #1@}
\newcommand{\tabincell}[2]{\begin{tabular}{@{}#1@{}}#2\end{tabular}}
\makeatother
\usepackage{multicol}
\usepackage{multirow}
\usepackage{booktabs}
\usepackage{bbding}
\usepackage{longtable}
\usepackage{supertabular}
\usepackage{pifont}
\usepackage{float}

\begin{document}

\title{Graph Reinforcement Learning Application to Co-operative Decision-Making in Mixed Autonomy Traffic: Framework, Survey, and Challenges}

\author{Qi Liu,~\IEEEmembership{Graduate Student Member,~IEEE}, Xueyuan Li$^{*}$, Zirui Li$^{*}$,~\IEEEmembership{Graduate Student Member,~IEEE}, Jingda Wu,~\IEEEmembership{Graduate Student Member,~IEEE}, Guodong Du, Xin Gao, Fan Yang, Shihua Yuan
\thanks{$^{*}$Corresponding authors: Xueyuan Li and Zirui Li}
\thanks{Qi Liu, Xueyuan Li, Zirui Li, Guodong Du, Xin Gao, Fan Yang, and Shihua Yuan are with the School of Mechanical Engineering, Beijing Institute of Technology, Beijing, China. (E-mails: 3120195257@bit.edu.cn;  lixueyuan@bit.edu.cn; 3120195255@bit.edu.cn; guodongdu\_robbie@163.com; 3120210298@bit.edu.cn; 3120225230@bit.edu.cn; yuanshihua@bit.edu.cn.)}
\thanks{Zirui Li is also with The Chair of Traffic Process Automation, "Friedrich List" Faculty of Transport and Traffic Sciences, TU Dresden, Germany}
\thanks{Jingda Wu is with the School of Mechanical and Aerospace Engineering, Nanyang Technological University, Singapore, 639798. (E-mail: jingda001@e.ntu.edu.sg)}
\thanks{Guodong Du is also with the Institute of Dynamic System and Control, ETH Zurich, Zurich, Switzerland.}
}

\markboth{Journal of \LaTeX\ Class Files,~Vol.~14, No.~8, August~2021}%
{Shell \MakeLowercase{\textit{et al.}}: A Sample Article Using IEEEtran.cls for IEEE Journals}

\maketitle

\begin{abstract}
Proper functioning of connected and automated vehicles (CAVs) is crucial for the safety and efficiency of future intelligent transport systems. Meanwhile, transitioning to fully autonomous driving requires a long period of mixed autonomy traffic, including both CAVs and human-driven vehicles. Thus, collaboration decision-making for CAVs is essential to generate appropriate driving behaviors to enhance the safety and efficiency of mixed autonomy traffic. In recent years, deep reinforcement learning (DRL) has been widely used in solving decision-making problems. However, the existing DRL-based methods have been mainly focused on solving the decision-making of a single CAV. Using the existing DRL-based methods in mixed autonomy traffic cannot accurately represent the mutual effects of vehicles and model dynamic traffic environments. To address these shortcomings, this article proposes a graph reinforcement learning (GRL) approach for multi-agent decision-making of CAVs in mixed autonomy traffic. First, a generic and modular GRL framework is designed. Then, a systematic review of DRL and GRL methods is presented, focusing on the problems addressed in recent research. Moreover, a comparative study on different GRL methods is further proposed based on the designed framework to verify the effectiveness of GRL methods. Results show that the GRL methods can well optimize the performance of multi-agent decision-making for CAVs in mixed autonomy traffic compared to the DRL methods. Finally, challenges and future research directions are summarized. This study can provide a valuable research reference for solving the multi-agent decision-making problems of CAVs in mixed autonomy traffic and can promote the implementation of GRL-based methods into intelligent transportation systems. The source code of our work can be found at \url{https://github.com/Jacklinkk/Graph_CAVs}.
\end{abstract}


\section{Introduction}
\IEEEPARstart{I}{ntelligent} transportation system plays an important role in both economic and social development, and connected and automated vehicles (CAVs) are an essential part of intelligent transportation systems \cite{7995802}. Before fully autonomous driving is achieved, CAVs will operate for a certain period in mixed autonomy traffic, which includes both CAVs and human-driven vehicles (HVs) \cite{zheng2020analyzing}. Therefore, the collaboration between CAVs and HVs and the communication between CAVs need to be carefully considered to ensure that CAVs can perform cooperative driving behaviors in mixed autonomy traffic \cite{hang2021cooperative}. Driving instructions of autonomous vehicles (AVs) are generated in decision-making systems. However, the simultaneous generation of driving instructions for multiple CAVs requires multi-agent decision-making systems. Therefore, designing a highly intelligent and reliable multi-agent decision-making system for CAVs is crucial to generate reasonable driving behaviors in mixed autonomy traffic, which could improve the efficiency and safety of future intelligent transportation systems \cite{Intro2}.

The detailed overview of decision-making technologies for AVs was presented in \cite{overview1,overview2}. These researches state that Reinforcement learning (RL) has been an effective method for solving decision-making problems because it can find optimal solutions in uncertain environments and does not require large labeled datasets. However, the dimensionality of the state and action space in mixed autonomy traffic is high. Therefore, applying the RL-based methods to mixed autonomy traffic applications face the problem of dimensional catastrophe, which significantly reduces the efficiency. To extend the RL-based methods to the high-dimensional state and action spaces, deep reinforcement learning (DRL)-based methods have been developed by embedding neural networks into the RL-based methods. In this way, the problems in complex and dynamic environments with high computational efficiency can be effectively handled without relying on prior knowledge. Therefore, the DRL-based methods have been widely applied to the decision-making process in mixed autonomy traffic \cite{yu2019distributed,palanisamy2020multi,ha2020leveraging,li2021reinforcement,han2022physics}. However, when solving a multi-agent decision-making problem of CAVs, a DRL-based method faces difficulties in modeling the mutual effect of vehicles in mixed autonomy traffic. This results in low cooperative behaviors, which can lead to danger or even traffic accidents. Thus, highly accurate representation methods for mixed autonomy traffic have been urgently needed. 

Graph representation is a powerful method to capture topological relationships and model the mutual effect of vehicles. However, a graph neural network (GNN) is needed to process the graphic features generated by the graph representation further. The combination of graph representation, GNN, and DRL-based methods can be used to design a graph reinforcement learning (GRL)-based method that can effectively solve the multi-agent decision-making problem of CAVs in mixed autonomy traffic. 

The main characteristics of the GRL-based methods can be summarized as follows: 1) Mixed autonomy traffic is modeled as a graph. Particularly, a vehicle is regarded as a node of the graph, while the mutual effects of vehicles are regarded as edges of the graph \cite{naderializadeh2020graph}. 2) A GNN is adopted for feature extraction; extracted features are fed to the policy network to generate the driving behaviors of CAVs. A number of studies have used GRL-based methods to generate cooperative behaviors. In \cite{jiang2018graph}, two multi-head attention graph convolutional layers were used for feature extraction, and behaviors for different agents were generated using deep q-learning (DQN). In \cite{chen2021graph}, GNN and DQN were combined (GCQ) for multi-agent cooperative controlling of CAVs in a merging scenario. The two mentioned studies have shown that the implementation of GRL can provide better results than using DRL in interactive scenarios. 

To verify preliminary the optimization effect advantage of the GRL-based methods relative to the DRL-based methods, this study conducts a simple ablation experiment first. A GRL-based model is trained in a highway ramping scenario: graph convolutional networks (GCN, a typical GNN method) \cite{kipf2016semi} with proximal policy optimization (PPO) method \cite{schulman2017proximal}. The proposed model is compared with the model without a GCN, and results show that the GRL-based method performs better than the DRL-based method. 


In summary, the GRL-based methods have great potential to improve the decision-making performance for CAVs in mixed autonomy traffic. Therefore, this paper focuses on the study of the GRL-based methods in mixed autonomy traffic, including the development of a modular GRL framework, the review of the relevant literature, and the comparative study of GRL-based methods. This research can lay a foundation for related research, help to understand, explore, and improve the GRL-based methods, and promote the development of these methods in the field of intelligent transportation systems. The main contributions of this article can be summarized as follows:

\begin{itemize}
    \item A generic and modular GRL framework for multi-agent decision-making of CAVs in mixed autonomy traffic is designed. The proposed framework contains four modules: mixed autonomy traffic module, graph representation module, GRL module, and driving behaviors module. The corresponding elements and functions are explained in detail;
    \item A systematic review of the DRL- and GRL-based methods for decision-making in mixed autonomy traffic is presented, focusing on the research topics that recent work has addressed;
    \item A comparative study of different GRL-based methods with an open-source code is conducted based on the designed framework. An ablation experiment is performed to verify the optimization effect advantage of the GRL-based methods compared to the corresponding DRL-based methods. Moreover, a comparison experiment is conducted to explore the performances of different GRL methods;
    \item Challenges and future research topics of the GRL-based methods for multi-agent decision-making in mixed autonomy traffic are discussed based on the current research status.
\end{itemize}

The rest of this paper is organized as follows. In Section \Rmnum{2}, a generic and modular GRL framework for multi-agent decision-making of CAVs in mixed autonomy traffic is introduced. In Section \Rmnum{3}, a comprehensive review of state-of-the-art DRL- and GRL-based methods is provided. In Section \Rmnum{4},  a comparative study of different GRL-based methods is conducted based on the designed framework. In Section \Rmnum{5}, challenges and future research topics are summarized and discussed. Finally, the main conclusions are drawn in Section \Rmnum{6}.

\section{Proposed Framework}
Before conducting detailed research, it is essential to introduce a framework of the GRL-based methods. This section presents a modular GRL framework for multi-agent decision-making in mixed autonomy traffic based on related research. The presented framework contains the following modules: mixed autonomy traffic module, graph representation module, GRL module (including GNN and DRL module), and driving behaviors module. The mixed autonomy traffic module is the basis of the proposed framework. The graph representation module is used to generate graphic features of mixed autonomy traffic and input them to the GRL module. The GRL module is the core of the framework to generate driving policies. The driving behavior module selects driving behavior according to the driving policies and inputs it to the mixed autonomy traffic to update the environment state.

\begin{figure*}[thpb]
  \centering
  \includegraphics[scale=0.29]{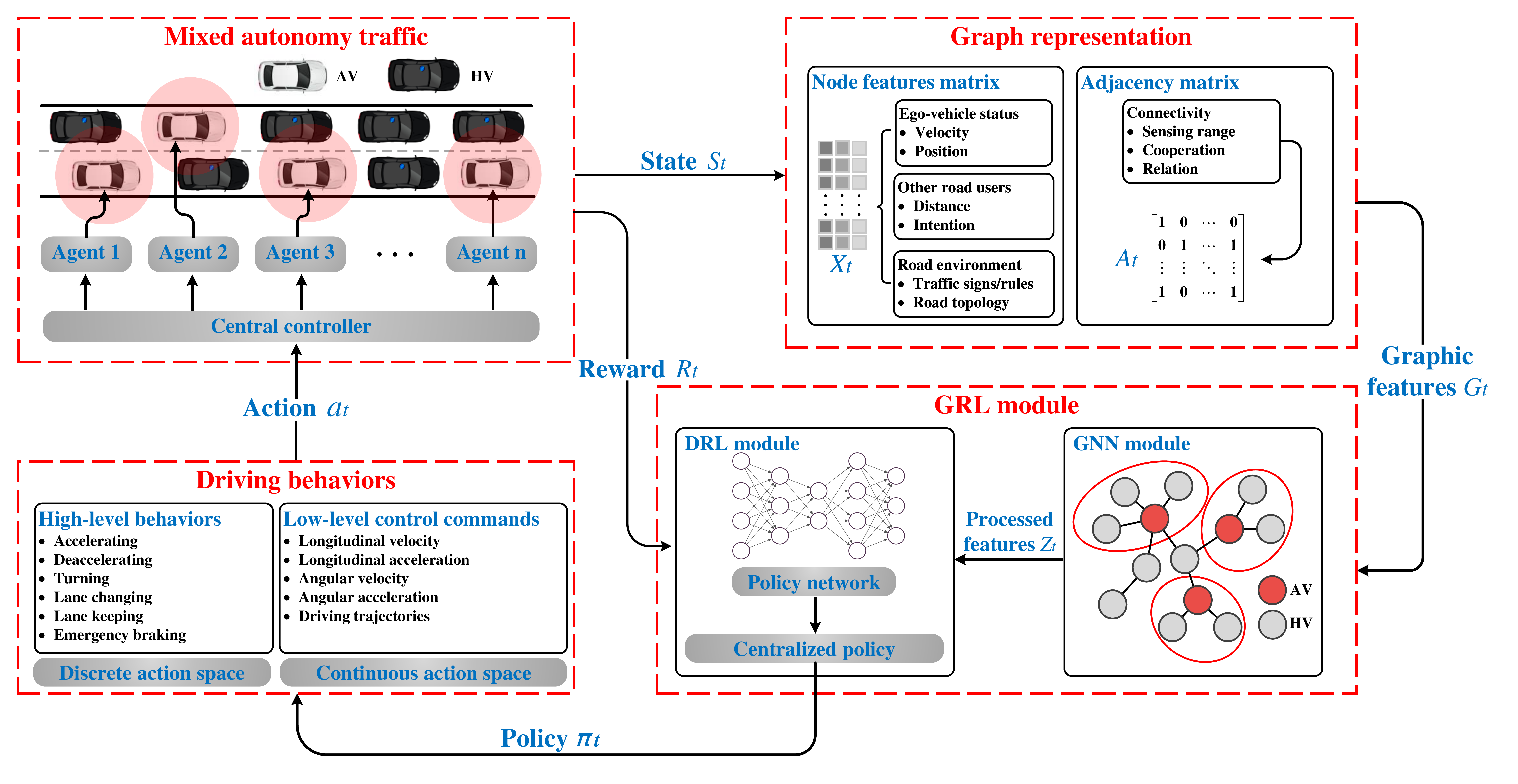}
  \caption{The schematic diagram of the proposed framework.}
  \label{figurelabel_1}
\end{figure*}

\subsection{GRL Framework Architecture}
The complete design of the proposed framework is illustrated in Fig. \ref{figurelabel_1}. The decision-making problem addressed by the proposed framework can be modeled as a finite horizon Markov decision process (MDP) \cite{bellman1957markovian}, which is defined by tuple \((S, a, P, R, \gamma)\). In tuple \((S, a, P, R, \gamma)\), \(S\) denotes a set of states that represent the current states of mixed autonomy traffic; \(a\) is a set of actions performed by the vehicles; \(P\) represents the transition probability distribution, \(R\) denotes a reward function used to evaluate the performance of actions taken by vehicles in the current states, \(\gamma\in(0,1]\) is a discount factor for future reward. Moreover, in many real-world application domains, it is not possible for each vehicle to observe all features of the current environment state. Thus, the decision-making problem can be further modeled as a partially-observable Markov decision process (POMDP) \cite{kaelbling1998planning}, which is defined by tuple \((S, a, P, R, \Omega, O, \gamma)\), where \(\Omega\) denotes the set of observations at the current state, and \(O\) is the observation model generating a possibility distribution over the possible observations. In the proposed GRL framework, both the MDP and the POMDP are implemented by neural networks, and the goal is to explore an optimal policy that maximizes the rewards by solving the decision-making model.

The mixed autonomy traffic is modeled as a graph, where a vehicle is regarded as a node of the graph, and the mutual effect of vehicles is regarded as edges of the graph. The graph is defined as \(G=\{N, E\}\), where \(N=\{n_{i}, i\in\{1,2,...,n\} \}\) is a set of node attributes and \(E=\{e_{ij}, i,j\in\{1,2,...,n\} \}\) is a set of edge attributes; \(n\) denotes the number of nodes in the constructed graph, and it is equal to the total number of vehicles. 

At a specific time step \(t\), the current state of mixed autonomy traffic \(S_{t}\) is extracted through graph representation to generate graphic features \(G_{t}\). The GNN module uses the graphic features as input and generates processed features \(Z_{t}\), which are then fed to the DRL module. Next, policy \(\pi_{t}\) is produced by the DRL module, and a set of actions \(a_{t}\) is generated to update the state of the traffic scenario state. Finally, the reward \(R_{t}\) of the current time step is fed back to the GRL module to update model parameters.

In the proposed modular framework, the graph representation methods can be adjusted according to the traffic scenario characteristics and modeling approaches. The GNN and DRL modules in the GRL module can be freely substituted according to the actual need to achieve different combinations. The proposed framework can also be adjusted to different traffic environments.

\subsection{Mixed Autonomy Traffic}
The mixed autonomy traffic considered in this study consists of CAVs and HVs, where CAVs are controlled by a GRL-based algorithm. The mixed autonomy traffic model can be constructed according to specific practical conditions.

The design of mixed autonomy traffic can refer to the scenarios constructed in some traffic simulation software. The Highway-env \cite{highway-env} provides many typical traffic scenarios, such as highways, intersections, and roundabouts. Flow \cite{wu2021flow} is a DRL-based framework for mixed autonomy traffic, which acts as an interface between traffic simulators (e.g. Sumo \cite{dlr127994} and Aimsun \cite{AimsunManual}) and RL libraries. The Flow framework not only provides typical traffic scenarios, but also creates several benchmarks for the development and verification of RL algorithms; it also supports the import operation of road network files (e.g., OpenStreetMap) to simulate traffic operations under real-world conditions.

\subsection{Problem Formulation}
The constructed GRL framework uses the graphic features extracted through graph representation as input. The graphic features are processed, and driving behaviors are generated and then fed to the controlled vehicles in the mixed autonomy traffic model.

\subsubsection{Graph Representation}
In the proposed framework, graph representation is used to extract the mixed autonomy traffic features and feed them to the GNN module. The graph representation consists of the node feature matrix \(N_{t}\) and the adjacency matrix \(A_{t}\), which are explained in the following.

\textbf{Node feature matrix:} The state of vehicles in a mixed traffic scenario is represented by the node feature matrix, whose elements are feature vectors of vehicles. The node feature matrix can be expressed as follows:

\begin{equation}
    N_{t}=[V_{t}^{1},V_{t}^{2},\cdots,V_{t}^{i},\cdots,V_{t}^{n}]^{T}
\end{equation}
where \(\{V_{t}^{i},i\in[1,n]\}\) denotes the feature vector of the \(i\)th vehicle. Feature vectors can contain multi-dimensional data on a vehicle, such as position, speed, and attitude.

\textbf{Adjacency matrix:} The mutual effect and interaction between vehicles are represented by the adjacency matrix, which can be represented as follows:

\begin{equation}
A_{t}=
\begin{bmatrix}
 e_{11}& e_{12}& \cdots&  & \cdots& e_{1n}\\ 
 e_{21}& e_{22}& \cdots&  & \cdots& e_{2n}\\ 
 \vdots& \vdots&  \ddots&  &  & \vdots\\ 
 & &  &  e_{ij}&  & \\ 
 \vdots& \vdots&  &  &  \ddots& \vdots\\ 
 e_{n1}& e_{n2}& \cdots&  & \cdots& e_{nn} 
\end{bmatrix}
\end{equation}
where \(\{e_{ij},i,j\in[1,n]\}\) denotes the edge value of the \(i\)th and \(j\)th vehicles; the edge value can be derived through the predefined interaction model of vehicles.

\textbf{Scenario classification:} The mixed autonomy traffic scenarios can be divided into open- and closed-loop traffic scenarios according to the invariance of the number of vehicles in a traffic scenario. In an open-loop traffic scenario (e.g., roundabout or ramping scenario), the number of vehicles changes; while in a close-loop traffic scenario (e.g., ring network or vehicle platoon), the number of vehicles is fixed. 

It should be noted that for an open-loop traffic scenario, the above node feature matrix and adjacency matrix cannot be directly input into the GRL-based model to generate driving policy. Namely, since vehicles are entering and exiting a given scenario, the number of observed vehicles in the considered road network area changes dynamically. However, in the graph representation process, the features of each vehicle in the considered environment need to be stored in the corresponding position in the feature matrix. In addition, in the action output process, actions executed by a vehicle are defined by elements in the corresponding position in the action matrix. Therefore, an index matrix is required to record the vehicles that currently exist in the open-loop scenario at each time step. Each vehicle is numbered and then recorded in the corresponding location of the index matrix. The index matrix is described as follows:

\begin{equation}
    I_{t}=[Veh_{1},Veh_{2},\cdots,Veh_{i},\cdots,Veh_{n}]
\end{equation}

\noindent where \(\{Veh_{i}, i\in[1,n]\}\) indicates the existence of each vehicle; if \(Veh_{i}=1\),  the \(i\textrm{th}\) vehicle exists in the current environment, otherwise \(Veh_{i}=0\). 

In a closed-loop traffic scenario, features of a vehicle are automatically assigned to a specific position in the feature matrix, and the actions executed by the vehicle are selected from the elements in the corresponding position in the action matrix. Thus, the node feature matrix and the adjacency matrix can be directly input into the GRL module.

Define \(\odot\) as the matrix operation at the corresponding position according to the index matrix. Then the graph representation at the current time step in an open-loop traffic scenario can be formulated as \(S_{t}=[N_{t},A_{t}]\odot I_{t}\); meanwhile, in a close-loop traffic scenario, the graph representation can be directly formulated as \(S_{t}=[N_{t},A_{t}]\). 

The node feature matrix and the adjacency matrix can be freely defined according to actual requirements, and different graph representation methods can be implemented into the proposed framework. Various methods for constructing the node feature matrix are summarized in TABLE \ref{tab21}, and methods for constructing the adjacency matrix are given in TABLE \ref{tab22}. 

\begin{table*}
\begin{center}
\caption{Summary of The Node Feature Matrix Constructing Methods.}
\label{tab21}
\begin{tabular}{cm{4.2cm}cccccccc}
\toprule
\multirow{3.5}{*}{Refs} & \multirow{3.5}{4.2cm}{\centering Method} & \multirow{3.5}{*}{Scenario} & \multicolumn{7}{c}{Components} \\
\cmidrule{4-10}
~ & ~ & ~ & Speed & Acceleration & Position & Gap & \tabincell{c}{Speed\\Difference} & Intention & Lane\\
\midrule

\cite{chen2021deep} & Features of the ego vehicle and its four neighboring vehicles in the sensing range. & Highway Merging & \checkmark & ~ & \checkmark & ~ & ~ & ~ & ~\\
\midrule

\cite{zhou2022multi} & Features of the ego vehicle and its four neighboring vehicles in the sensing range. & Lane-changing & \checkmark & ~ & \checkmark & ~ & ~ & ~ & ~\\
\midrule

\cite{chen2021graph} & Feature vector of all vehicles (HVs and AVs). & Highway ramping & \checkmark & ~ & \checkmark & ~ & ~ & \checkmark & \checkmark\\
\midrule

\cite{li2021reinforcement} & Feature vector of the whole vehicle platoon. & Vehicle platoon control &
\checkmark & ~ & ~ & \checkmark & \checkmark & ~ & ~\\
\midrule

\cite{shi2020efficient} & Feature vector of ego vehicle and neighboring vehicles within the sensing range. & Various scenarios & \checkmark & ~ & \checkmark & \checkmark & \checkmark & ~ & ~\\
\midrule

\cite{hart2020graph} & Feature vector of all vehicles (HVs and AVs). & Lane-changing &
\checkmark & ~ & \checkmark & ~ & ~ & ~ & ~\\
\midrule

\cite{vinitsky2018benchmarks} & Feature vector of all vehicles (HVs and AVs). & Various scenarios &
\checkmark & ~ & \checkmark & \checkmark & ~ & ~ & ~\\
\midrule

\cite{xu2018reinforcement} & Information on the ego vehicle and other vehicles. & Lane-changing & \checkmark & ~ & ~ & \checkmark & ~ & ~ & \checkmark\\
\midrule

\cite{devailly2021ig} & Feature vector of all vehicles in the sensing range. & Traffic signal control &
\checkmark & ~ & \checkmark & ~ & ~ & ~ & \checkmark\\
\midrule

\cite{yu2019distributed} & Feature vector of the ego vehicle and neighboring vehicles within the sensing range. & Highway cruising &
\checkmark & ~ & \checkmark & \checkmark & ~ & ~ & \checkmark\\
\midrule

\cite{bouton2019cooperation} &  Feature vector of all vehicles (HVs and AVs). & Highway merging &
\checkmark & \checkmark & \checkmark & ~ & ~ & \checkmark & ~\\
\midrule

\cite{bai2022hybrid} & Feature vector of the ego vehicle and neighboring vehicles within the sensing range & Highway cruising. & 
\checkmark & \checkmark & ~ & \checkmark & ~ & \checkmark & ~\\
\bottomrule

\end{tabular}
\end{center}
\end{table*}

\begin{table*}
\begin{center}
\caption{Summary of The Adjacency Matrix Constructing Methods.}
\label{tab22}
\begin{tabular}{cccc}
\toprule
Refs & Methods & Scenario & Components\\
\midrule

\cite{chen2021graph} & Graph convolutional network (GCN) & Highway ramping & Information sharing between vehicles.\\
\midrule

\cite{shi2020efficient} & Graph attention network (GAT) & Various scenarios & Gaussian speed field using the Gaussian process regression (GPR) model.\\
\midrule

\cite{hart2020graph} & A directed graph with a self defined GNN. & Highway lane-changing & Relative distance between vehicles.\\

\bottomrule
\end{tabular}
\end{center}
\end{table*}

\subsubsection{Driving Behaviors}
According to the output level, driving behaviors can be separated into two categories: high-level behaviors and low-level control commands \cite{yu2020hierarchical}. High-level behaviors mostly include merging, overtaking, and lane keeping, whereas low-level control commands include velocity and acceleration in various vehicle control directions. Driving behaviors are represented as an action space, which can be divided into discrete action space and continuous action space. High-level behaviors can only be represented as a discrete action space; whereas low-level control commands can be represented as a discrete action space. Different policy-generated methods of the DRL module generate different action spaces, which in turn generate different categories of driving behavior.

The discrete action space is composed of a finite number of actions, which is typically the entire set of action commands available for a given task. For instance, in a lane-change scenario, the discrete action space can be defined as \(a=[change\ to\ left,\ go\ straight,\ change\ to\ right]\). The discrete action space is encoded using one-hot vectors, where each encoded point corresponds to an action command, and all encodings are mutually incompatible.

The continuous action space consists of specific values of control commands. For instance, in a highway scenario, the continuous action space can be defined as \(a=[a_{t}, \theta_{t}]\), where \(a_{t}\) denotes the longitudinal acceleration and \(\theta_{t}\) denotes the steering angle. The continuous action space is encoded using multi-dimensional (or one-dimensional) vector, where each encoded position represents a control command. The control commands are normally limited to a certain value range, and specific values of the control commands are determined based on the adopted control strategy. The continuous action space can be discretized at a certain granularity, but in this case, the trade-off between the control accuracy and the action space dimension has to be considered.

As the output data of the proposed GRL framework, the available levels of driving behavior have to correspond to the categories of policy-generated methods of the DRL module. The correspondence between driving behaviors and the DRL methods is described in detail in Section.\ref{GRL}. 

\subsection{GRL} \label{GRL}
The GRL module contains two sub-modules: the GNN module and the DRL module. This model uses graphic features as input and outputs policy as the basis for action selection.

\subsubsection{GNN Module}
GNN is a neural network type for processing data represented by a graph data structure \cite{sanchez2021gentle}. The GNN module uses graphic features extracted by graph representation as input data and outputs processing features that are then fed to the DRL module for further processing. The general calculating process of the GNN module can be defined by:

\begin{equation}
    G_{t}=\Phi_{\textrm{GNN}}(S_{t})
\end{equation}
where \(G_{t}\) denotes the graphic features processed by the GNN module; \(\Phi_{\textrm{GCN}}\) denotes the graph convolution operator.

Different GNN methods can be implemented into the proposed framework. In this work, studies that can help to select an appropriate GNN method are mentioned. The original concept of GNN was proposed in \cite{scarselli2008graph}. The overviews of various GNN methods and applications were presented in \cite{zhou2020graph,he2021overview}. A comprehensive survey of GNN was provided in \cite{wu2020comprehensive}, as well as the open-source scripts, benchmark datasets, and model evaluation of GNNs.

\subsubsection{DRL Module}
DRL incorporates deep learning into RL, allowing agents to make decisions based on unstructured input data without manual engineering of the state space \cite{8585411}. In the proposed framework, the DRL module uses features processed by the GNN module as input data and generates policy for action selection as output data. The driving policy can be defined as follows:

\begin{equation}
    \pi_{t}=\Phi_{\textrm{DRL}}(G_{t})
\end{equation}
where \(\pi_{t}\) is the driving policy generated by the DRL module; \(\Phi_{\textrm{DRL}}\) denotes the DRL model.

Different types of DRL methods generate different driving policy categories. The DRL methods can be divided into value- and policy-based methods. Value-based methods are applicable only to discrete action spaces; these methods aim to generate the driving policy consisting of values of different actions and then select driving behaviors according to the value of each available action. The policy-based methods are applicable to both discrete and continuous action spaces. They can generate both stochastic deterministic driving policies, and then driving behaviors are selected accordingly. The calculation methods of driving behaviors using different DRL methods are summarized in TABLE \ref{tab1}.

\begin{table}
\begin{center}
\caption{Calculation Methods of Driving Behaviors using Different DRL Methods.}
\label{tab1}
\begin{tabular}{ccm{3.8cm}}
\toprule
Categories & \tabincell{c}{Driving\\policy} & \multirow{1}{3.8cm}{\centering Derivation of driving behaviors}\\

\midrule
\multirow{1}*{Value-based} & \tabincell{c}{State-value\\function} & \textbf{Discrete}: Value of each available action.\\
\midrule

\multirow{6}*{\tabincell{c}{Policy-based}} & \tabincell{c}{Deterministic\\policy} & \textbf{Continuous}: Specific numerical instruction of each action. \\
\cmidrule{2-3}

~ & \multirow{3.5}*{\tabincell{c}{Stochastic\\policy}} & \textbf{Discrete}: Probability of each available action.\\
\cmidrule{3-3}
~ & ~ & \textbf{Continuous}: Normal distribution of each available action.\\

\bottomrule
\end{tabular}
\end{center}
\end{table}

Different DRL methods can be implemented into the proposed framework. In this work, a comprehensive summary of studies that present the DRL methods was provided. An overview of fundamentals, typical algorithms, applications, and resources of DRL was provided in \cite{li2017deep}. A survey of DRL for autonomous driving was presented in \cite{kiran2021deep}, while a survey of DRL for intelligent transportation systems is given in \cite{haydari2020deep}

\section{Review of Decision-Making Algorithm in Mixed Autonomy Traffic}
\begin{figure*}[thpb]
  \centering
  \includegraphics[scale=0.32]{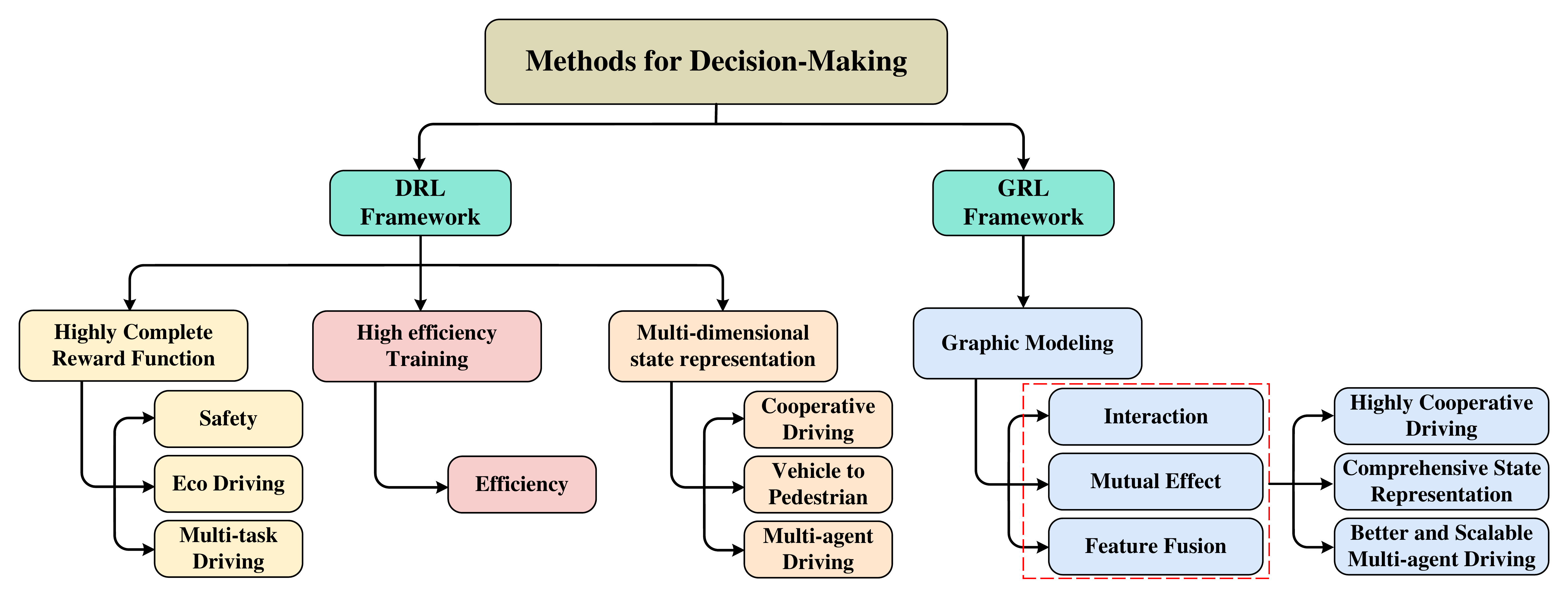}
  \caption{The taxonomy of research problems addressed by the DRL- and GRL-based methods for decision-making in the mixed autonomy traffic.}
  \label{figureTaxnomy}
\end{figure*}

This section presents a review of the decision-making algorithm in mixed autonomy traffic. Considering that the DRL-based methods denote a crucial part of the GRL-based framework, the research on the DRL can be transferred to the GRL, which can significantly contribute to the GRL algorithm development. Therefore, both the DRL- and GRL-based approaches are reviewed in this section. In sub-section A, the state-of-the-art DRL-based approaches are summarized in detail, focusing on the issues that recent research has addressed. In sub-section B, the advantages of the GRL-based methods over the DRL-based methods are emphasized, and several GRL-based methods for decision-making in mixed autonomy traffic are presented. The taxonomy of research problems addressed by the DRL- and GRL-based methods for decision-making in the mixed autonomy traffic is depicted in Fig. \ref{figureTaxnomy}. 

\subsection{DRL Methods}
Typical RL-based methods can solve decision-making problems in a limited state and action spaces. However, the RL-based methods suffer from problems of high dimensionality and low solution efficiency in complex driving scenarios. To overcome these problems, a deep neural network is incorporated into the RL methods to construct a DRL-based framework. In decision-making for AVs, the DRL-based methods have a strong ability to handle complex and dynamic driving environments and can efficiently produce near-optimal actions based solely on interactions with the environment without any prior knowledge about the underlying system \cite{frikha2021reinforcement,boute2021deep}. Numerous studies have focused on the DRL-based methods to solve the decision-making problem in mixed autonomy traffic. In this work, the state-of-art literature is reviewed. Several typical DRL-based approaches are presented in TABLE \ref{tabDRL}.

\subsubsection{Safety}
Safety is the first priority in cooperative decision-making. Learning how to drive safely is essential for CAVs in mixed autonomy traffic.

A primary possible solution for designing a safe policy to define additional restrictions on action selection. In \cite{mirchevska2018high}, the DQN was combined with formal safety verification to ensure that only safe actions could be selected, and highly desired velocity was reached with nearly no collision. However, the trade-off between safety and efficiency should be further considered. In \cite{bernhard2019addressing}, a risk-sensitive approach was proposed in the T-intersection scenario; offline distributional DQN was used to solve the model, and an online risk assessment was performed to evaluate the probability distribution of the generated actions. The results showed that the collision rate was less than 3\%. Similarly, in \cite{bouton2019safe}, a "model-checker" based safety RL method was proposed to guarantee the safety of intersections in complex environments. A recurrent neural network was trained to generate beliefs, and driving instructions were generated based on the DQN and according to the constraints of the safety threshold. Approximately 100 steps were necessary to complete the goal for the given scenario at a low collision rate. In \cite{schmidt2021can}, a safe decision-tree policy was designed to ensure safe distance in a highway overtaking scenario; collision was obviously reduced in randomized initialization. However, the reward function needed further development because the overall reward decreased when collisions were reduced.

Another possible solution for ensuring driving safety is constructing a safe reward function to train a DRL-based model to generate safe driving behaviors. In \cite{kamran2020risk}, a strict risk-based reward function was derived to punish risk situations instead of only collision-related ones. A generic “risk-aware DQN” was proposed for generating safe behaviors. Results showed that a success rate of near 95\% could be achieved under a low training vibration. Moreover, in \cite{hu2022roadside}, an assessment module based on the Bayesian inference was designed to ensure safe reward generation. In \cite{alizadeh2019automated}, the trade-off between safety and agility was considered when designing the reward function; overall braking induced by the lane-changing behaviors was mainly minimized while encouraging the AV to speed up.

Several other techniques have also been adopted to improve driving safety. In \cite{hoel2020reinforcement}, multiple neural networks were assembled with additional randomized prior functions to optimize the typical DQN capacity. In this way, safe driving could be realized in more uncertain intersections. Results showed that the success rate of more than 95\% could be achieved under a collision rate of less than 5\% . However, the constructed simulation environment was too simple. In \cite{kuutti2021arc}, adversarial robust control (ARC) was implemented in a highway leader-follower driving scenario. The A3C was selected as a basic framework, and a protagonist network was constructed to control the follow vehicle, whereas an adversary network was constructed to control the lead vehicle. The number of collisions decreased by 90.25\%. Nevertheless, the non-lead-follower scenario should be also considered. In \cite{seong2021learning}, an attention mechanism was introduced to focus on more spatially and temporally important environmental features to generate self-attention features. This enabled safe and efficient driving decisions even under noisy sensory data, and a success rate of more than 87\% was achieved at a low collision rate and average braking time.

\subsubsection{Efficiency}
Another critical research topic in decision-making is how to ensure high efficiency. In this study, efficiency refers to solve the DRL model of decision-making with high real-time performance, which is critical to practical applications in CAVs.

In \cite{yavas2020new}, the rainbow DQN was combined with a safely-driving rewarding scheme to achieve high sample efficiency. The trained model converged to stable reward after only 200k training steps compared with baseline (1M training steps). In \cite{qiao2018pomdp}, decision-making at an intersection was modeled as hierarchical-option MDP (HOMDP), where only the current observation was considered instead of the observation sequence over a time interval to reduce the computational cost. A success rate of more than 97\% was achieved, and 50\% lesser number of steps were needed to finish the driving task compared with the baseline. In \cite{liu2019learning}, human demonstration with the supervised loss was implemented into the training of a double DQN for a better exploration strategy to boost the learning process. A success rate of over 90\% could be reached with only 100 training epochs.

\subsubsection{Eco-driving}
Eco-driving can reduce resource waste and have significant economic benefits. Learning how to control the ego vehicle more efficiently and cooperate with other vehicles to improve transportation efficiency could be beneficial to energy saving.

Vehicle platoon control has been a hot topic because improper driving behavior of a vehicle can adversely affect the driving efficiency of other vehicles. In \cite{prathiba2021hybrid}, a hybrid DRL and genetic algorithm for smart-platooning (DRG-SP) was proposed. A genetic algorithm was implemented into the DRL-based framework to overcome the slow convergence problem and ensure long-term performance. The driving policy was updated through a rank-based replay memory to make highly optimal decisions. Results showed that the energy consumption was reduced by 8.57\% while maintaining high efficiency. In \cite{li2021reinforcement}, a communication proximal policy optimization (CommPPO) was proposed for eco-driving. A predecessor–leader-follower typology in the platoon was utilized with a new reward communication channel to guarantee efficient information transmission and avoid the lazy-agent problem. In addition, curriculum learning was first adopted to train a small-size platoon to facilitate the training process of the whole vehicle platoon. Results showed that fuel consumption was reduced by 11.6\%.

Several other driving scenarios have also been investigated in the research on eco-driving. In \cite{lichtle2021fuel}, an "I-210 network" was designed. Multi-agent PPO with a traffic smoothing controller was proposed to eliminate traffic shockwaves. The designed system achieved a 25\% fuel consumption reduction at a 10\% penetration rate. However, only two vehicles were controlled in the constructed scenario. In \cite{liu2021efficient}, an efﬁcient on-ramp merging strategy (ORMS) was proposed. The D3QN was combined with prioritized experience replay to learn the lane-changing behaviors, and a motion planning algorithm based on time-energy optimal control was developed by adding time term into the reward function to generate an optimal trajectory. Results showed that the fuel economy and traffic efficiency could be improved by 43.5\% and 41.2\%. In \cite{bai2022hybrid}, a unity-based simulator was developed, and a mixed traffic intersection scenario was designed. A hybrid RL (HRL)-based framework, which combined the rule- and DRL-based modules was proposed for eco-driving at intersections. Particularly, the rule-based module was used to ensure good collaboration between the two types of strategies, while a dueling DQN was implemented into the DRL module to generate driving behaviors by capturing both visual and logical information. Results showed that energy consumption and travel time were reduced by 12.70\% and 11.75\%, respectively.

\subsubsection{Cooperative driving}
In this study, cooperative driving mainly refers to the decision-making of a single CAV considering collaboration with other HVs. Learning how to perform cooperative driving behavior in mixed autonomy traffic has significant implications for improving traffic efficiency.

The highly comprehensive modeling of interactions has great potential to improve cooperation between vehicles. In \cite{bouton2019cooperation}, HVs were modeled with different cooperation levels in the DRL framework. Typical DQN was combined with a belief updater to generate driving instructions under different cooperation levels. The number of time-out failures was obviously reduced compared with the baseline. Moreover, in \cite{wang2021harmonious}, a multi-agent RL method for harmonious lane-changing was developed. The proposed harmonious driving method relied only on the ego vehicles’ limited sensing results to balance the overall and individual efﬁciencies. In addition, a reward function that combined individual efﬁciency with the overall efﬁciency for harmony was designed. Results showed that a high mean vehicle flow rate could be reached under congested conditions.

Better prediction of other vehicles' behaviors can help to generate cooperative behaviors of the ego vehicle. In \cite{kamran2021high}, the Deep-Sets DQN was proposed to handle the dynamic number of vehicles. The proposed model can efficiently predict cooperative drivers’ behaviors based on their historical data and generate high-level cooperative instructions; the MPC was used to generate driving trajectories. Similarly in \cite{el2021novel}, a high accuracy data-driven model was developed based on a directed graphic model to predict the intention of HVs. The predicted results were then input into the DRL framework to generate cooperative driving behaviors. Results showed that an average speed of 31.8m/s could be reached with stable speed deviation.

\subsubsection{Vehicle-to-pedestrian interaction}
Apart from the cooperation between vehicles, vehicle-pedestrian interaction is also important for safe autonomous driving.

One solution for ensuring pedestrian safety is to generate vehicle braking commands directly. In \cite{chae2017autonomous}, an autonomous braking system based on a DQN was designed. The output of the system was a series of braking commands of different strengths. The collision rate reaches zero when the time-to-collision (TTC) interval was longer than 1.5s. However, only a single vehicle and person were considered in the constructed scenario. In \cite{deshpande2021navigation}, a multi-objective reward function was designed in the DQN framework for navigation in urban environments in the presence of pedestrians. Both acceleration and braking commands were generated, and the results indicate that both driving safety and efficiency were optimized. 

Predicting pedestrians' behaviors could contribute to the safe driving of autonomous vehicles. In \cite{papini2021reinforcement}, a safe speed network was constructed and integrated with the DRL agent. Moreover, a risk assessment was performed to predict the behaviors of distracted pedestrians. The predicted results were then input into the double DQN with an integrated safe speed network to generate driving behaviors.

Complete modeling of pedestrians' intentions is important since pedestrians' behaviors are highly uncertain. In \cite{trumpp2022modeling}, a pedestrian was modeled as a DRL agent to define the vehicle-pedestrian interaction as a multi-agent DRL problem. Two levels of the pedestrian models and vehicle models were established. The obtained results indicated that a collision rate of 0.135\% was achieved under maximal noisy level, and the DRL pedestrian model could learn an intelligent crossing behavior. However, whether modeling the pedestrian as a DRL agent could reduce the requirement for vehicle intelligence should be further explored.

\subsubsection{Multi-agent}
In general, multiple CAVs need to be controlled in mixed autonomy traffic. Thus, multi-agent decision-making technology is highly demanded.

In \cite{schester2021automated}, a simple multi-agent DRL framework was proposed to solve the problem of a highway merging scenario. The acceleration command of each AV was generated using the status of other vehicles as input data. Collision-free performance was achieved at an on-ramp length of 70m or longer with vehicles that were 5m or more apart. However, only two vehicles (one driving on the main lane, another driving on the merge lane) were controlled in the constructed scenario; still additional vehicles should also be further considered. More vehicles were considered in \cite{9565000}, where the REINFORCE algorithm was used to generate driving behaviors based on local observations for an arbitrary number of controlled vehicles. Results showed that the near-optimal throughput with 33\%-50\% controlled vehicles could be achieved.

A more complete multi-agent decision-making system was designed in \cite{zhou2022multi}, and a multi-agent A2C method with the parameter-sharing mechanism and multi-objective reward function was proposed to achieve decentralized control of multiple AVs. Feature vectors of the ego vehicle and its neighboring vehicle were used as input data, and driving instructions of all AVs were then generated. Moreover, the designed reward function was used to evaluate the performance of every single AV, and the transition of each vehicle was stored into experience replay individually. Then, the experience replay was sampled for model training. The authors conducted similar research in \cite{chen2021deep}, where the main improvement was that a priority-based safety supervisor was developed to avoid invalid behaviors to reduce collision numbers.

Modeling the interaction between different vehicles can provide more reasonable driving behaviors for each vehicle. In \cite{yu2019distributed}, a dynamic coordination graph was proposed to model the continuously changing topology during vehicles’ interactions. A tubular Q-learning was proposed to generate driving behaviors. In addition, two mechanisms (the global coordination mechanism and the local coordination mechanism) were employed to extend the approach to more general and complex situations with any number of vehicles. Results indicated good performance in scenarios with different numbers of vehicles.

\subsubsection{Multi-task driving}
Driving efficiency of CAVs can be further improved by optimizing multiple driving tasks simultaneously.

A straightforward approach for optimizing multiple driving tasks simultaneously is to establish multi-objective reward functions to train an AV to execute multiple driving tasks simultaneously. In \cite{kai2020multi}, a unified four- dimensional vectorized reward function was derived and combined with a DQN to solve the navigation problem at different types of intersections. The designed reward function consisted of the reward values generated by four different driving actions in the current state. In \cite{9304647}, two objectives, collision avoidance for safety and jerk minimization for passenger comfort, were investigated in designing the reward function. The DDPG was used for behavior generation, and results showed that vehicle jerk is reduced by 73\% with nearly no collision in the highway merging scenario. Similarly in \cite{wang2021multi}, driving speed and fuel efficiency were jointly considered in designing the reward function. The AC algorithm, which takes the visual image as input and outputs the control commands to achieve the end-to-end driving, was used. However, the verification scenario included only the ego vehicle but no other vehicles.

In \cite{he2020multi}, more types of objects were implemented into the reward function. Safety, comfort, economy, and transport efficiency were considered in design a multi-mode reward function. The PPO was employed, and results indicated that a feasible and effective driving policy for autonomous electric vehicles was achieved. However, more combinations of weight coefficients need to be investigated. In addition in \cite{ye2021meta}, similar objects were considered in the design of the reward function. The main difference was that meta RL was adopted to improve the generalization capability of the DRL model for more complex environments. The overall success rate was up to 20\% higher than the benchmark model, and the collision rate was reduced by 18\%.

Decoupling the driving tasks into several sub-task is another possible solution for dealing with multi-task driving. In \cite{triest2020learning}, the driving tasks were modeled through a hierarchical framework integrating high-level policy and low-level control. High-level driving behaviors were generated by the A2C and then input into the vehicle kinematic model to generate acceleration and steering angle commands. Results showed that the collision rate was less than 5\%. In \cite{gangopadhyay2021hierarchical}, the driving tasks were decomposed into several simple tasks, and a hierarchical program-triggered RL-based (HPRL) framework was established to train different agents to complete the decomposed sub-tasks simultaneously. The proposed method demonstrated good training efficiency in multi-task autonomous driving.

\subsubsection{Other}
Some other research objectives have also been considered in recent studies. In \cite{de2020ethical}, driving ethics were considered, including three different policies, particularly, Rawlsian contractarianism, utilitarianism, and egalitarianism. A search-based method was used to generate ethical driving instructions. In \cite{pusse2019hybrid}, the benchmark establishment process was mainly studied, and an OpenDS-CTS benchmark based on the major German in-depth road accident study GIDAS was proposed to verify safe decision-making in vehicle-pedestrian accident scenarios. Moreover, a hybrid method named HyLEAP, which combines belief tree and DRL, was proposed to generate collision-free behaviors.

\subsection{GRL Methods}
\begin{figure}[thpb]
  \centering
  \includegraphics[scale=0.24]{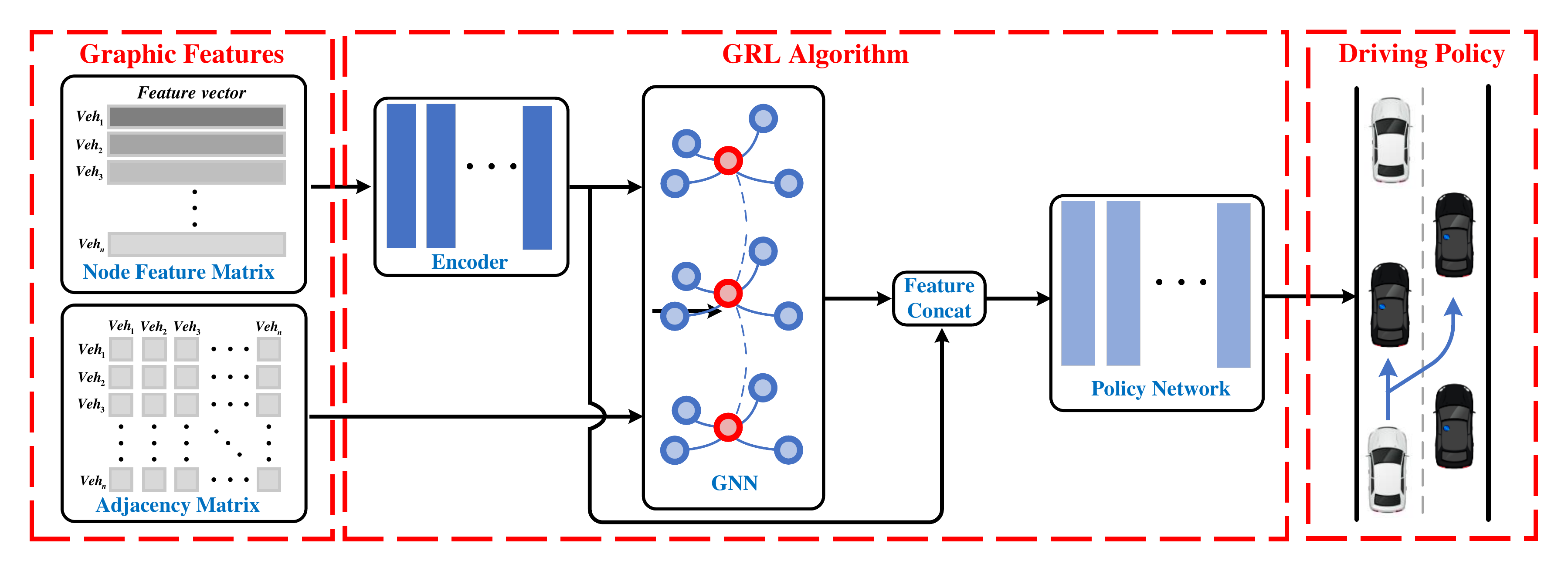}
  \caption{The detailed structure of the GRL algorithm.}
  \label{figureGRLStructure}
\end{figure}

The DRL-based methods are prevalent for decision-making in mixed autonomy traffic. However, when employing only DRL to solve multi-agent decision-making and cooperative driving, system complexity increases significantly, and it is difficult to model relationships between agents. Since a GNN can obtain the topological relationship and facilitate the modeling of the mutual effects of multiple agents, it has great potential to improve decision-making performance in mixed autonomy traffic. This section summarizes the relative research on the GRL-based methods. The detailed structure of a GRL-based method is illustrated in Fig. \ref{figureGRLStructure}. The overview of the GRL-based approaches is given in TABLE \ref{tabGRL}.

A straightforward solution has been to model the mixed autonomy traffic as a graph, representing features of vehicles as a node feature matrix and mutual effects between vehicles as an adjacency matrix. Therefore, a GNN can be used to aggregate the above two matrices into a DRL-based framework to generate driving behaviors. In \cite{chen2021graph}, a highway ramping scenario was constructed and modeled as an undirected graph. The GCN was used to acquire the data collected through collaborative sensing, while cooperative lane-changing decisions were generated by the DQN. The results showed that the average reward was higher than those obtained by the rule-based and LSTM methods in different traffic densities . However, the generated behaviors did not correspond to the current vehicles. Based on \cite{chen2021graph}, two improvement solutions were proposed. In \cite{s22134935}, a generalized single-agent GRL training method was developed. The training results were applied to multi-agent training to reduce the computational cost. However, continuous action space should be considered for generating acceleration commands. In \cite{s22124586}, a multi-mode reward function with a decision-weighted coefficient matrix was derived to train multiple decision-making modes  in different traffic scenarios. Four decision-making strategies, including aggressive incentive (AGGI), aggressive punishment (AGGP), conservative incentive (CONI), and conservative punishment (CONP), were trained with a multi-step double DQN. Results showed that higher reward and average speed could be achieved.

Exploring additional ways of modeling interactions between vehicles is significant to improving the effectiveness of the GRL-based methods. In \cite{hart2020graph}, a highway lane-changing scenario was modeled as a directed graph, and graph representation was implemented based on the relative position between vehicles. Further, in \cite{klimke2022cooperative}, an intersection scenario was constructed, and the connection between vehicles was modeled based on their turning intentions. In \cite{shi2020efficient}, an attention mechanism was introduced to capture the mutual interplay between vehicles to achieve better cooperative control. Moreover, a dynamic adjacency matrix based on the Gaussian speed field using the Gaussian process regression (GPR) model was constructed to capture spatial and temporal interactions between surrounding vehicles. A graph attention network (GAT) was used for graphic feature extraction, while the PPO was employed for policy generation. Various scenarios were verified, and results indicated a higher average reward than that of the baseline.

Another method is to use a GNN to fuse multiple feature categories without modeling the mixed autonomy traffic as a graph. In \cite{cai2021dignet}, various traffic scenarios were designed in the Carla simulator. Graph node features of vehicles and bird-eye view images were concatenated and input in the GAT. Then, the aggregated features were fused with the motion vector and route of the ego vehicle and fed to a multi-layer perceptron (MLP) model to generate throttle and steering commands. Safe navigation in a complex driving environment was achieved while satisfying traffic rules. Similar research was conducted in \cite{cai2022dq}. The main difference was that only graph node features and bird-eye view were fused and input in the GAT. The D3QN was combined with a noisy network to improve policy exploration and generation. Results showed that a success rate of over 96\% was achieved in training scenarios.

\begin{table*}
\caption{Summary of Exemplary DRL-based Approaches for Decision-making in Mixed Autonomy Traffic.}
\label{tabDRL}
\begin{tabular}{ccccm{1.5cm}m{2.5cm}ccm{3.4cm}}
\toprule
\multirow{3}{*}{\tabincell{c}{Task\\solved}} & \multirow{3}{*}{Refs} & \multirow{3}{*}{\centering Methods} & \multirow{3}{*}{Scenario} & \multirow{3}{1.5cm}{\centering Verification} & \multirow{3}{2.5cm}{\centering Performance} & \multicolumn{3}{c}{\centering Characteristics}\\

\cmidrule{7-9}
~ & ~ & ~ & ~ & ~ & ~ & \tabincell{c}{Main\\solution} & \tabincell{c}{Multi-\\agent} & \multirow{1}{3.4cm}{\centering Remarks}\\
\midrule

\multirow{8}{*}{Safety} &\cite{bernhard2019addressing} & \tabincell{c}{Distribu-\\tional DQN} & Intersection & \centering Numerical simulation & Collision rate of less than 3\%. & 
Safe policy & \ding{53} & An online risk assessment mechanism is introduced to evaluate the probability distribution of different actions.\\
\cmidrule{2-9}

~ & \cite{kamran2020risk} & \tabincell{c}{Risk-aware\\DQN} & Intersection & \centering Simulation in Carla & More than 95\% success rate with steady performance. & \tabincell{c}{Safe reward\\function} & \ding{53} & A stricter risk-based reward function is constructed to solve the model.\\
\cmidrule{2-9}

~ & \cite{kuutti2021arc} & SAC & \tabincell{c}{Various\\scenarios} & \centering Simulation in Carla & Success rate of more than 87\% with a low collision rate. & \tabincell{c}{Attention\\mechanism} & \ding{53} & An attention-based spatial-temporal fusion driving policy is proposed.\\
\midrule

\multirow{5}{*}{\tabincell{c}{High\\efficiency\\solving}} & \cite{qiao2018pomdp} & DQN & Intersection & \centering Simulation in SUMO & Over 97\% success rate with a small total number of finishing steps. & \tabincell{c}{Hierarchical\\framework} & \ding{53} & Hierarchical Options MDP (HOMDP) is utilized to model the scenario.\\
\cmidrule{2-9}

~ & \cite{liu2019learning} & \tabincell{c}{Double\\DQN} & \tabincell{c}{Highway\\lane-changing} & \centering Numerical simulation & Over 90\% success rate is achieved with only 100 training epochs. & Demonstration & \ding{53} & Human demonstration with supervised loss is introduced.\\
\midrule

\multirow{5}{*}{\tabincell{c}{Eco\\driving}} & \cite{li2021reinforcement} & PPO & \tabincell{c}{Vehicle\\platoon} & \centering Simulation in SUMO & Fuel consumption is reduced by 11.6\%. & \tabincell{c}{Oscillation\\resuction} & \ding{53} & The predecessor–leader follower typology is proposed.\\
\cmidrule{2-9}

~ & \cite{bai2022hybrid} & \tabincell{c}{Dueling\\DQN} & Intersection & \centering Unity Engine & Energy consumption is reduced by 12.70\%. & \tabincell{c}{Hybrid\\framework} & \ding{53} & The rule-based strategy and the DRL strategy are combined.\\
\midrule

\multirow{5}{*}{\tabincell{c}{Coopera-\\tive driving}} & \cite{wang2021harmonious} & DQN & \tabincell{c}{Highway\\lane-changing} & \centering Numerical simulation & 6529 mean vehicle flow rate in congested conditions. & \tabincell{c}{Behavior\\prediction} & \checkmark & Individual efﬁciency with overall efﬁciency for harmony is combined.\\
\cmidrule{2-9}

~ & \cite{kamran2021high} & \tabincell{c}{Deep-Sets\\DQN} & \tabincell{c}{Highway\\merging} & \centering Numerical simulation & Low comfort cost is achieved under cooperative driving. & \tabincell{c}{Behavior\\prediction} & \ding{53} & Cooperative drivers from their vehicle state history are identified.\\
\midrule

\multirow{5}{*}{\tabincell{c}{Vehicle to\\Pedestrian}} & \cite{chae2017autonomous} & DQN & \tabincell{c}{Pedestrian\\crossing} & \centering Simulation in PreScan & Collision rate reaches zero when TTC is higher than 1.5s. & \tabincell{c}{Brake Control} & \ding{53} & An autonomous braking system is designed with different braking strengths.\\
\cmidrule{2-9}

~ & \cite{papini2021reinforcement} & \tabincell{c}{Double\\DQN} & \tabincell{c}{Distracted\\pedestrian\\ crossing} & \centering Simulation in OpenDS & Different safe speed range is verified under various pedestrian situations. & \tabincell{c}{Behavior\\prediction} & \ding{53} & A risk assessment is performed to predict the behaviors of pedestrians.\\
\midrule

\multirow{7}{*}{\tabincell{c}{Multi-agent\\driving}} & \cite{schester2021automated} & DDPG & \tabincell{c}{Highway\\merging} & \centering Numerical simulation & Collision-free performance is achieved at the merging ramp. & \tabincell{c}{Parameter\\sharing} & \checkmark & Collision avoidance is emphasized in the interaction between vehicles.\\
\cmidrule{2-9}

~ & \cite{chen2021deep} & \tabincell{c}{Improved\\A2C} & \tabincell{c}{Highway\\merging} & \centering Simulation in Highway-env & Zero collision rate is achieved in three tested modes. & \tabincell{c}{Parameter\\sharing} & \checkmark & A priority-based safety supervisor is developed to reduce collision.\\
\cmidrule{2-9}

~ & \cite{yu2019distributed} & \tabincell{c}{Tubular\\Q-learning} & \tabincell{c}{Highway\\cruising} & \centering Graphical simulation & High average reward with good lane-keeping behaviors. & \tabincell{c}{Interaction\\modeling} & \checkmark & A dynamic coordination graph is proposed to model the interactive topology.\\
\midrule

\multirow{10}{*}{\tabincell{c}{Multi-task\\driving}} & \cite{kai2020multi} & \tabincell{c}{Multi-\\task DQN} & Intersection & \centering Simulation in SUMO & Success rate is higher than 87\%. & \tabincell{c}{multi-objective\\reward function} & \ding{53} & Multiple tasks are represented by a unified four-dimensional vector with a vectorized reward function.\\
\cmidrule{2-9}

~ & \cite{9304647} & \tabincell{c}{DDPG} & \tabincell{c}{Highway\\merging} & \centering Simulation in SUMO & Vehicle jerk is reduced by 73\% with nearly no collision. & \tabincell{c}{multi-objective\\reward function} & \ding{53} & Collision avoidance for safety and jerk minimization for passenger comfort are both investigated.\\
\cmidrule{2-9}

~ & \cite{gangopadhyay2021hierarchical} & \tabincell{c}{DQN\textbackslash DDPG} & \tabincell{c}{Various\\scenarios} & \centering Simulation in Carla & 100\% success rate with no traffic rule violation. & \tabincell{c}{Tasks\\decoupling} & \ding{53} & Multiple agents are trained with the different simple task under the hierarchical DRL framework.\\

\bottomrule
\end{tabular}
\end{table*}

\begin{table*}
\begin{center}
\caption{Summary of the GRL-based Approaches for Decision-making in Mixed Autonomy Traffic.}
\label{tabGRL}
\begin{tabular}{cccm{1.5cm}m{3.1cm}ccm{3.7cm}}
\toprule
\multirow{2}{*}{Refs} & \multirow{2}{*}{\centering Methods} & \multirow{2}{*}{Scenario} & \multirow{2}{1.5cm}{\centering Verification} & \multirow{2}{3.1cm}{\centering Performance} & \multicolumn{3}{c}{\centering Characteristics}\\

\cmidrule{6-8}
~ & ~ & ~ & ~ & ~ & \tabincell{c}{Main solution} & \tabincell{c}{Multi-agent} & \multirow{1}{3.7cm}{\centering Remarks}\\
\midrule

\cite{chen2021graph} & \tabincell{c}{GCN+DQN} & \tabincell{c}{Highway\\ramping} & \centering Simulation in SUMO & Better than those of the rule-based and LSTM at different traffic density values. & \tabincell{c}{Graph\\modeling} & \checkmark & The traffic scenario is modeled as an undirected graph. However, the generated behaviors don’t correspond to the current vehicles.\\
\midrule

\cite{s22134935} & \tabincell{c}{GCN+DQN} & \tabincell{c}{Highway\\ramping} & \centering Simulation in SUMO & The network convergence and training efficiency are improved. & \tabincell{c}{Graph\\modeling} & \checkmark & A generalized single-agent GRL training method is proposed and extended to the multi-agent framework.\\
\midrule

\cite{s22124586} & \tabincell{c}{GCN+DQN} & \tabincell{c}{Highway\\ramping} & \centering Simulation in SUMO & High reward and average speed can be achieved. & \tabincell{c}{Graph\\modeling} & \checkmark & A multi-mode reward function with a decision-weighted coefficient matrix is derived to achieve the training of multiple decision-making modes.\\
\midrule

\cite{hart2020graph} & \tabincell{c}{Directed\\graph+PPO} & \tabincell{c}{Highway\\lane-changing} & \centering Numerical simulation & 81.6\% success rate is achieved at 11.1\% collision rate. & \tabincell{c}{Graph\\modeling} & \checkmark & Graph representation is implemented based on the relative position between vehicles.\\
\midrule

\cite{klimke2022cooperative} & \tabincell{c}{GCN+TD3} & \tabincell{c}{Intersection} & \centering Simulation in Highway-env & Flow rate in the intersection is significantly improved. & \tabincell{c}{Graph\\modeling} & \checkmark & The varying number of vehicles in the scenario is handled by a flexible graph representation.\\
\midrule

\cite{shi2020efficient} & \tabincell{c}{GAT+PPO} & \tabincell{c}{Various\\scenarios} & \centering Simulation in SUMO & Average reward is increased in all the tested scenarios. & \tabincell{c}{Graph\\modeling} & \checkmark & The attention mechanism is introduced to capture mutual interplay among vehicles to achieve better cooperative control.\\
\midrule

\cite{cai2021dignet} & \tabincell{c}{DiGNet} & \tabincell{c}{Various\\scenarios} & \centering Simulation in Carla & Safe navigation in a complex driving environment while obeying traffic rules. & \tabincell{c}{Graphical\\feature fusion} & \ding{53} & Graph representation is fused with bird’s-eye views of the driving scenario and route information.\\
\midrule

\cite{cai2022dq} & \tabincell{c}{GAT+D3QN} & \tabincell{c}{Various\\scenarios} & \centering Simulation in Carla & Over 96\% success rate in the training scenarios. & \tabincell{c}{Graphical\\feature fusion} & \ding{53} & Graph representation is fused with bird’s-eye views. PID controller is implemented in decision-making module.\\

\bottomrule
\end{tabular}
\end{center}
\end{table*}

\section{Comparative Study of GRL-Based methods}

This study conducted a comparative study of different GRL-based algorithms on the same benchmark to investigate the utility of the GRL-based approach. Compared with the DRL-based methods, the GRL-based methods model the interaction between vehicles and have great potential to improve decision-making performance. Therefore, a modular GRL library based on the proposed framework was constructed to explore the effects of various GRL-based algorithms and the corresponding DRL-based algorithms systematically on the same traffic benchmark. The modular GRL library was designed using Python language and different third-party libraries. Simulations were performed using the Flow library \cite{wu2021flow} and SUMO platform \cite{dlr127994}. The main goal was to provide a valuable resource and a foundation for future relevant research. 

Compared to the existing GRL-based approaches for decision-making in mixed autonomy traffic, the main contributions of this research could be summarized as follows: 1) A modular GRL library was developed based on the proposed framework. According to particular requirements for a traffic environment, the graph representation methods, GNN methods, and DRL methods can be adjusted; 2) Comparison experiments with state-of-the-art methods were performed. The experimental results were analyzed comprehensively and systematically.

\subsection{Traffic Scenario Construction }
In the comparative study, two traffic scenarios were constructed for the purpose of validation of different GRL-based methods, namely a highway ramping scenario and a Figure-Eight scenario. The two scenarios were highly discriminated, the invariance of the number of vehicles and the type of action space were both different. In this way, the results of a comparative study based on these two scenarios were well represented. The two traffic scenarios are described in detail in the following.

\subsubsection{Highway Ramping Scenario}
\begin{figure}[h]
  \centering
  \includegraphics[scale=0.37]{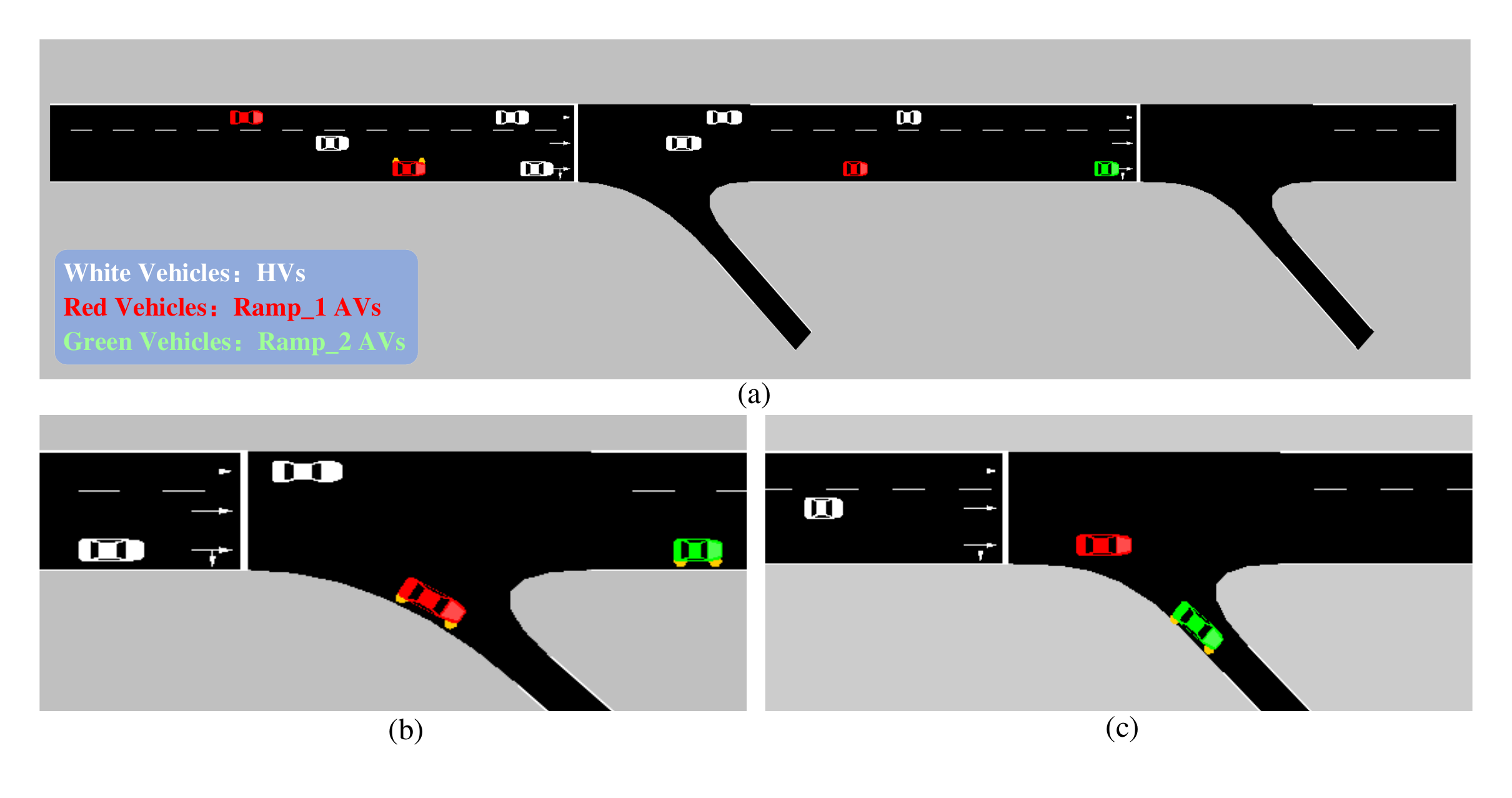}
  \caption{The highway ramping scenario: (a) the entire view of the traffic scenario; (b) the view of Ramp 1 where red AVs drive out; (c) the view of Ramp 2 where green AVs drive out.}
  \label{figurelabel_2}
\end{figure}

The highway ramping scenario was constructed based on \cite{chen2021graph,wu2021flow}. The scenario consisted of a three-lane highway with two ramp exits. It included HVs and two types of AVs. Different vehicles had different driving tasks, but they needed to cooperate to complete the scheduled driving tasks more efficiently. The illustration of the highway ramping scenario is given in Fig. \ref{figurelabel_2}. where white vehicles represent HVs, entering from the left side of the highway and exiting on the right side;  colored vehicles represent AVs that enter from the left side of the highway and exit at different ramps. Specifically, red vehicles denote Ramp\_1 AVs, merging at Ramp 1; green vehicles denote Ramp\_2 AVs, merging at Ramp 2.

\textbf{Graph Representation:} The highway ramping scenario was an open-loop scenario. In addition to the node feature matrix and the adjacency matrix, the scenario included the index matrix. Thus, the graph representation consisted of the node features matrix, adjacency matrix and index matrix.

The node features matrix \(N_{t}\in \mathbf{R}^{(m+n)\times 8}\) contained
feature vectors of all vehicles. The feature vector was defined as follows:

\begin{equation}
    V_{i} = [v_{i},x_{i},l_{i},t_{i}]
\end{equation}
where \(V_{i}\) represents the feature vector of the \(i\textrm{th}\) vehicle. \(v_{i}=v_{i\_actual}/v_{max}\) denotes the normalized longitudinal speed of vehicles relative to the maximum limit of longitudinal speed; \(x_{i}=x_{i\_longitudinal}/L_{highway}\) denotes the normalized longitudinal coordinate of vehicles relative to the length of highway; \(l_{i}\) denotes the one-hot encoding matrix of current lane position (left-lane, right-lane and middle-lane) of vehicles; \(t_{i}\) denotes the one-hot encoding matrix of the current intention of vehicles (e.g., change to left, go straight, and change to right). 

The adjacency matrix \(A_{t}\in \mathbf{R}^{(m+n)\times (m+n)}\) represents the information sharing between vehicles in the highway ramping scenario. The edge value in the adjacency matrix was calculated based on the following assumption: 1) All vehicles were connected with themselves; 2) AVs could communicate with each other. 3) AVs could communicated with the surrounding HVs in their sensing range. If the \(i\)th and \(j\)th vehicles share information, the edge value \(e_{ij}=1\); otherwise, \(e_{ij}=0\).

The index matrix was derived based on the existence of AVs and HVs in the current environment, and the formulation is as follows:

\begin{equation}
    I_{t}=[HV_{1},HV_{2},\cdots,HV_{m},AV_{1},AV_{2},\cdots,AV_{n}]
\end{equation}

\noindent where \(m\) denotes the total number of HVs, \(n\) denotes the total number of AVs. \(\{HV_{i}, i\in\{1,2,...,m\}\}\) indicates the existence of HVs, and if \(HV_{i}=1\), the \(i\textrm{th}\) HV exists in the current environment, otherwise \(HV_{i}=0\). \(\{AV_{i}, i\in\{1,2,...n\}\}\) indicates the existence of AVs; when \(AV_{i}=1\), the \(i\textrm{th}\) AV exists in the current environment, otherwise \(AV_{i}=0\).

\textbf{Driving Behaviors:} 
In a highway ramping scenario, the output driving behavior represented a list of high-level lane-change commands used to control the lateral motion of AVs. The longitudinal control of both HVs and AVs was achieved by the intelligent driver model (IDM) \cite{treiber2013traffic}, while the lateral control of HVs was achieved by the LC2013 lane-change model \cite{erdmann2015sumo}.

The driving behavior was characterized as a discrete action space. At each time step, the action space included different lane change instructions, which can be described as follows:

\begin{equation}
    a=[change\ to\ left,\ go\ straight,\ change\ to\ right]
\end{equation}

\begin{figure}[thpb]
  \centering
  \includegraphics[scale=0.28]{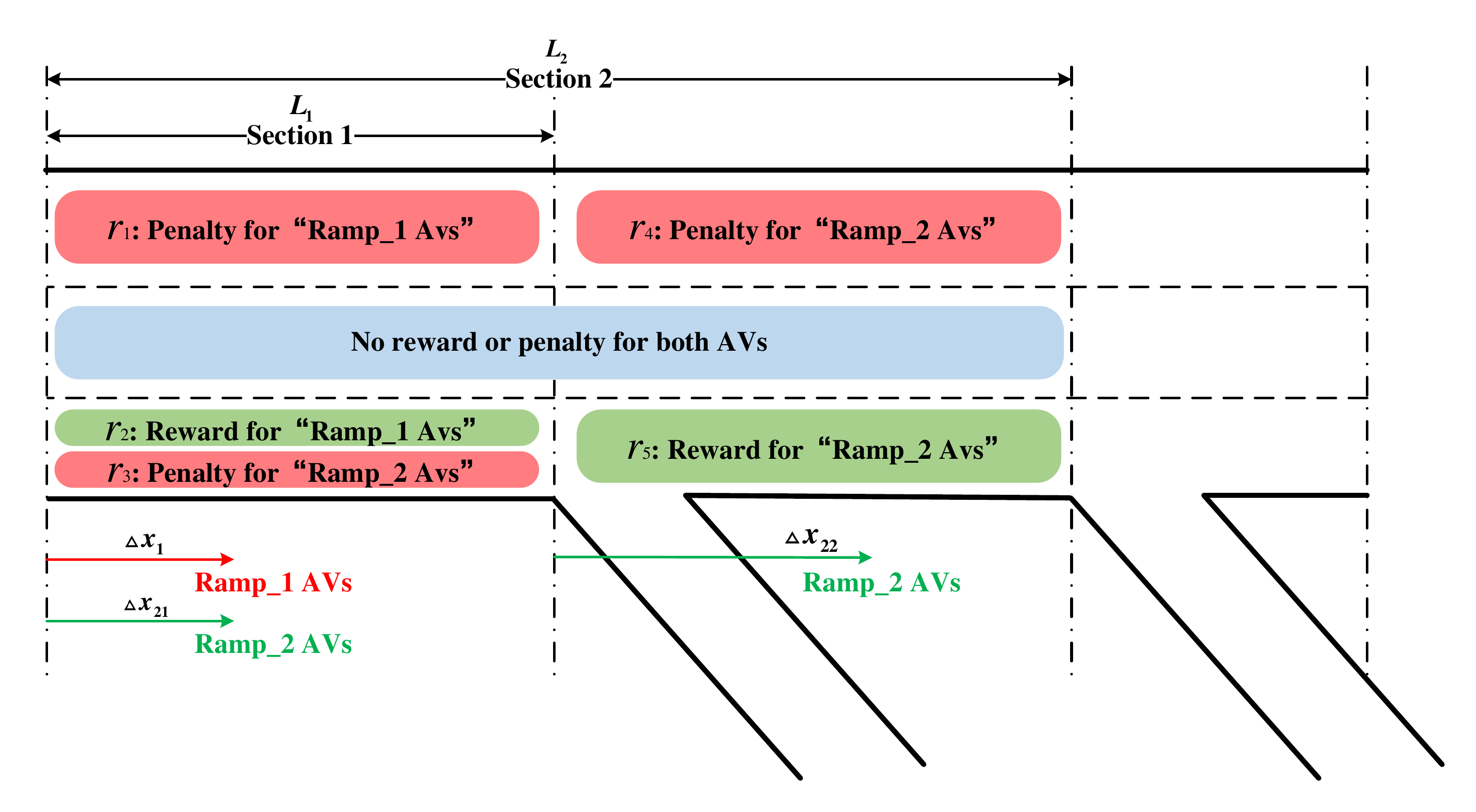}
  \caption{The illustration of the intention rewards on different road sections.}
  \label{figurelabel_4}
\end{figure}

\textbf{Reward Function:} 
The reward function was designed based on \cite{chen2021graph}. In the highway ramping scenario, the goal of AVs was to exit the corresponding ramp efficiently and safely while minimizing the impact on HVs. The reward function consisted of four parts: average speed reward, intention reward, lane- changing penalty, and collision penalty. The reward function was defined as follows:

\begin{equation}
    R=w_{1}R_{\textrm{I}}+w_{2}R_{\textrm{AS}}+w_{3}P_{\textrm{LC}}+w_{4}P_{\textrm{C}}
\end{equation}

\noindent where \(w_{i}\) denotes the weight of each item in the formulation.

\(R_{\textrm{I}}\) represents the intention reward to guarantee AVs can merge out from their prescribed ramp. The calculation of intention reward with road sections. The schematic diagram of the road sections segmentation is shown in Fig. \ref{figurelabel_4}, and the derivation of intention reward on different road section is presented in TABLE \ref{tab3}.

\begin{table}
\begin{center}
\caption{Derivation of Intention Rewards on Different Road Sections.}
\label{tab3}
\begin{tabular}{cm{4.5cm}c}
\toprule
\tabincell{c}{Reward\\type} & \centering Description & Derivation\\

\midrule
\(r_{1}\) &  Penalty for Ramp\_1 AVs driving on the leftmost lane decreases from 0 to -1 when approaching Ramp 1. & \(r_{1}=-\frac{\Delta x_{1}}{L_{1}}\) \\

\(r_{2}\) & Reward for Ramp\_1 AVs driving on the rightmost lane, decreases from 1 to 0 when approaching Ramp 1. & \(r_{2}=1-\frac{\Delta x_{1}}{L_{1}}\) \\

\(r_{3}\) & Penalty for Ramp\_2 AVs driving on the rightmost lane, decreases from 0 to -1 when approaching Ramp 1. & \(r_{3}=-\frac{\Delta x_{21}}{L_{1}}\) \\

\(r_{4}\) & Penalty for Ramp\_2 AVs driving on the leftmost lane, decreases from 0 to -1 when approaching Ramp 2. & \(r_{4}=-\frac{\Delta x_{22}}{L_{2}-L_{1}}\) \\

\(r_{5}\) & Reward for Ramp\_2 AVs driving on the rightmost lane, decreases from 1 to 0 when approaching Ramp 2. & \(r_{5}=1-\frac{\Delta x_{22}}{L_{2}-L_{1}}\) \\

\bottomrule
\end{tabular}
\end{center}
\end{table}

\(R_{\textrm{AS}}\) represents the average speed reward to encourage the driving behaviors performed by AVs to increase the system efficiency, which is obtained by:

\begin{equation}
    R_{\textrm{AS}}=\frac{1}{n}\sum\limits_{k=1}\limits^n\frac{V_{AV}^{k}}{V_{max}}
\end{equation}

\(P_{\textrm{LC}}\) denotes the lane-changing penalty to encourage AVs exit from the corresponding ramp with the minimum lane-changing operations; \(P_{\textrm{LC}}\) is the lane-changing penalty and \(P_{\textrm{LC}}=n_{LC}\), where \(n_{LC}\) is the the number of lane changes of AVs at the current time step.

\(P_{\textrm{C}}\) denotes the collision penalty to ensure system safety, and \(P_{\textrm{C}}=n_{col}\), where \(n_{col}\) is the number of collision at the current time step.

\textbf{Parameters Setting:} 
The parameters of the highway ramping scenario are given in TABLE \ref{tab4}.

\begin{table}[h]
\caption{Parameters Setting of the Highway Ramping Scenario.}
\label{tab4}
\begin{center}
\begin{tabular}{ccc}
\toprule
Parameters & Symbols & Value\\
\midrule 
Number of HVs & \(m\) & 6\\
Number of AVs & \(n\) & 6\\
Number of Ramp\_1 AVs & \(n_{\textrm{r1}}\) & 3\\
Number of Ramp\_2 AVs & \(n_{\textrm{r2}}\) & 3\\
Highway length & \(L\) & 200m\\
Longitudinal position of Ramp 1 & \(L_{\textrm{1}}\) & 80m\\
Longitudinal position of Ramp 2 & \(L_{\textrm{2}}\) & 160m\\
Speed limit of HVs & \(V_\textrm{{maxHVs}}\) & 60km/h\\
Speed limit of AVs & \(V_\textrm{{maxAVs}}\) & 75km/h\\
Inflow of HVs & \(P_\textrm{{HVs}}\) & 0.5veh/s\\
Inflow of AVs & \(P_\textrm{{AVs}}\) & 0.3veh/s\\
Longitudinal control of HVs & - & IDM\\
Longitudinal control of AVs & - & IDM\\
Lateral control of HVs & - & LC2013\\
Lateral control of AVs & - & GRL model\\
\bottomrule
\end{tabular}
\end{center}
\end{table}

\subsubsection{Figure-Eight Scenario}
\begin{figure}[thpb]
  \centering
  \includegraphics[scale=0.4]{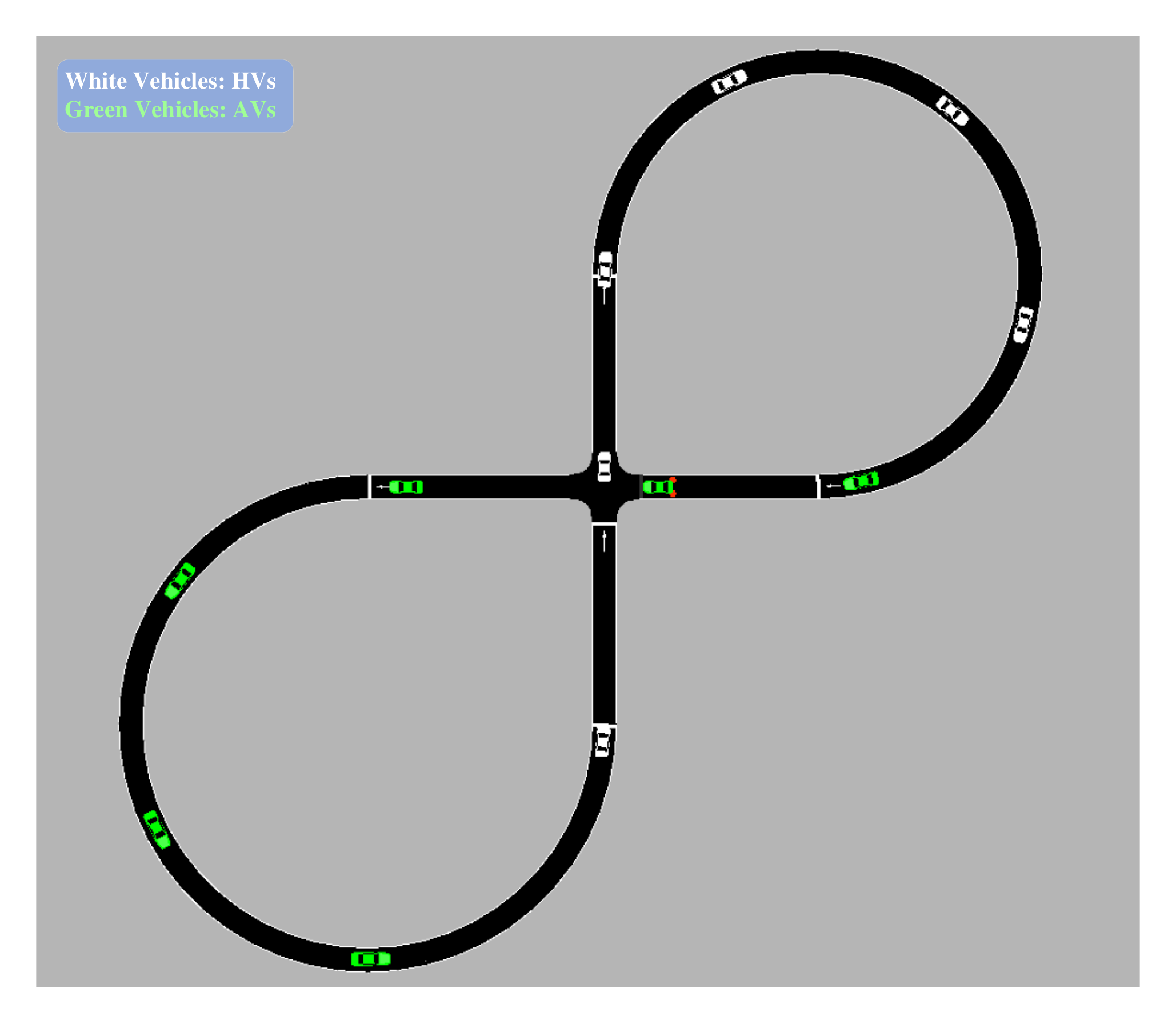}
  \caption{The Figure-Eight scenario.}
  \label{figurelabel_3}
\end{figure}

The Figure-Eight scenario was constructed based on \cite{wu2017emergent,wu2021flow}. The scenario acted as a closed representation of an intersection consisting of two single-lane ring networks. When vehicles arrived simultaneously at the intersection, they had to slow down to obey the right-of-way rule. This reduced the average speed of vehicles in the network. In this scenario, cooperative driving was demanded to increase the average speed of vehicles while ensuring safety to optimize intersection capacity. The illustration of the Figure-Eight scenario is shown in Fig. \ref{figurelabel_3}, white vehicles represent HVs, and green vehicles represent AVs. 

\textbf{Graph Representation:} The Figure-Eight scenario was a closed-loop scenario, so the index matrix was not required. The adjacency matrix for the Figure-Eight scenario was derived in the same way as for the highway ramping scenario. The node features matrix \(N_{t}\in \mathbf{R}^{(m+n)\times 2}\) was different, and the feature vector of each vehicle was presented as follows:

\begin{equation}
    V_{i} = [v_{i},x_{i}]
\end{equation}
where \(v_{i}=v_{i\_actual}/v_{max}\) denotes the normalized longitudinal speed of vehicles relative to the maximum limit of the longitudinal speed; \(X_{i}\) denotes the longitudinal coordinate of vehicles relative to the pre-specified starting point (i.e., the intersection point of the two ring networks).

\textbf{Driving Behaviors:} 
In the Figure-Eight scenario, the output driving behavior was a list of low-level control commands used to control the longitudinal motion of AVs, while the longitudinal control of HVs was achieved by the IDM.

The driving behavior was characterized as continuous action space. At each time step, the action space consisted of longitudinal acceleration that can be defined as \(a \in [a_{min},a_{max}]\), where \(a_{min}\) denotes the minimum longitudinal acceleration, and \(a_{max}\) denotes the maximum longitudinal acceleration.

\textbf{Reward Function:} 
The reward function was designed based on \cite{vinitsky2018benchmarks}. In the presented comparative study, the GNN was integrated into several state-of-art DRL methods to investigate the effectiveness of the GRL-based methods in decision-making in mixed autonomy traffic.

\begin{equation}
R=\textrm{max}(\|V_{d}\cdot \mathbf{1}^{(m+n)}\|_{2}-\|V_{d}- V\|_{2},0)/\|V_{d}\cdot \mathbf{1}^{(m+n)}\|_{2}
\end{equation}
\noindent where \(V_{d}\) is an desired large velocity to facility high speed, \(V\in \mathbf{R}^{m+n}\) is the velocity matrix of all vehicles in the current environment.

\textbf{Parameters Setting:} 
The parameters of the Figure-Eight scenario are summarized in TABLE \ref{tab5}.

\begin{table}[h]
\caption{Parameters Setting of the Figure-Eight Scenario.}
\label{tab5}
\begin{center}
\begin{tabular}{ccc}
\toprule
Parameters & Symbols & Value\\
\midrule 
Number of HVs & \(m\) & 6\\
Number of AVs & \(n\) & 6\\
Radius of the ring & \(r\) & 30m\\
Speed limit of vehicles & \(V_\textrm{max}\) & 100km/h\\
Desired speed of vehicles & \(V_\textrm{d}\) & 140km/h\\
Maximum longitudinal acceleration & \(a_\textrm{max}\) & 3m/s\textsuperscript{2}\\
Minimum longitudinal acceleration & \(a_\textrm{min}\) & -3m/s\textsuperscript{2}\\
Longitudinal control of HVs & - & IDM\\
Longitudinal control of AVs & - & GRL model\\
\bottomrule
\end{tabular}
\end{center}
\end{table}

\subsection{GRL Methods}
The GRL methods consist of the GNN method and the DRL method. In the proposed comparative study, the GNN was integrated with several state-of-art DRL methods to investigate the effectiveness of GRL methods for decision-making in mixed autonomy traffic. In addition, the common hyperparameters of the implemented GRL methods were the same to ensure the control variables for the comparative study.

\subsubsection{GNN Methods}
The difference between the GRL-based methods and the DRL-based methods relates to the implementation of graph representation and GNNs. The GCN \cite{kipf2016semi} was used in the GNN module to process the graphic features of the constructed traffic scenarios. The key of the GCN is the graph convolution process that can be expressed by:

\begin{equation}
    Z_{t}=\Phi_{\textrm{GCN}}(N_{t}, A_{t})=\sigma(D_{t}^{\frac{1}{2}}A_{t}D_{t}^{-\frac{1}{2}}N_{t}W_{t}+b)
\end{equation}

\noindent where \(Z_{t}\) denotes the graph convolutional features processed by the GCN; \(\phi_{\textrm{GCN}}\) denotes the graph convolution operator; \(D_{t}\) is computed based on \(A_{t}\), specifically, \(D_{ii}=\Sigma_{j}A_{ij}\); \(W_{t}\) is a layer-specific trainable weight matrix; \(b\) denotes the offset, and \(b=0\) in this research; \(\sigma\) denotes an activation function, and in this work, the ReLU \cite{glorot2011deep} function was selected since it could better fit the characteristics of biological neurons in neural network. 

\subsubsection{DRL Methods}
In the comparative study, various state-of-the-art DRL methods were used. As mentioned before, the existing DRL-based methods can be categorized into value- and policy-based methods. The value-based methods include the DQN and its different expansions, such as double DQN, dueling DQN, noisy DQN, DQN with prioritized replay buffer (PRE), distributional DQN, and rainbow DQN. The policy-based methods include REINFORCE, actor-critic (AC), Advantage actor-critic (A2C), proximal policy optimization (PPO), normalized advantage function and its extension (NAF, Double NAF), deep deterministic policy gradient (DDPG), twin delayed deep deterministic policy gradient (TD3), and soft actor-critic (SAC). Due to the space limitation, this study does not present all the listed method in detail, but more information about them can be found in the provided citations. The DRL methods implemented in the comparative study are summarized in TABLE \ref{tab6}. 

\begin{table*}
\begin{center}
\caption{List of DRL Methods Used in the Comparative Study.}
\label{tab6}
\begin{tabular}{ccccp{9.2cm}}
\toprule 
\multirow{1}{*}{Category} & \multirow{1}{*}{Algorithm} & \multirow{1}{*}{Refs} & Available scenario & \multirow{1}{10cm}{\centering Characteristic} \\ 

\midrule
\multirow{26}*{\tabincell{c}{Value- \\ Based}} & DQN & \cite{mnih2013playing} & Highway ramping & 
\tabincell{l}{1. Neural network is introduced to approximate action value function. \\2. A target Q network is used to generate and update the target Q-values \\ 3. Experience replay is applied to breakdown the correlation between samples.\\ 4. Suitable for large state spaces. \\ 5. Suffering from overestimation of action values under certain conditions.}\\
\cmidrule{2-5}

~ & Double DQN & \cite{van2016deep} & Highway ramping &
\tabincell{l}{\tabincell{l}{1. The selection of the action is decoupled from the evaluation of the \\ \quad target Q value.}\\ 2. Significantly mitigate bias caused by bootstrapping.}\\
\cmidrule{2-5}

~ & Dueling DQN & \cite{wang2016dueling} & Highway ramping &
\tabincell{l}{\tabincell{l}{1. The action value function is decoupled into state value function and \\ \quad optimal advantage function.}\\ 2. More accurate estimation of action value functions. \\ 3. The convergence process is accelerated. \\ 4. Overestimation of action value still exists.}\\
\cmidrule{2-5}

~ & Noisy DQN & \cite{fortunato2017noisy} & Highway ramping & \tabincell{l}{1. Add noisy function to the parameters of the neural network.\\ 2. Strong robustness. \\ 3. Provide a broader space for action exploration. \\ 4. Large computational cost.}\\
\cmidrule{2-5}

~ & DQN with PER & \cite{schaul2015prioritized} & Highway ramping &
\tabincell{l}{1. Assign different priorities to the samples in the replay buffer.\\ 2. High sample utilization.
\\ 3. High learning efficiency. \\ 4. The learning rate needs to be adjusted reasonably.}\\
\cmidrule{2-5}

~ & Distributional DQN & \cite{bellemare2017distributional} & Highway ramping & \tabincell{l}{1. The DRL formulation is modeled from a distributional perspective.\\ 2. The histogram is chosen to represent the estimate of the value distribution.
\\ 3. More accurate risk assessment of different actions.}\\
\cmidrule{2-5}

~ & Rainbow DQN & \cite{hessel2018rainbow} & Highway ramping &
\tabincell{l}{1. Integrate all the previous DQN-based methods.\\ 2. Multi-steps learning is utilized to accelerate learning speed.}\\
\midrule

\multirow{36}*{\tabincell{c}{Policy- \\ Based}} & REINFORCE & \cite{williams1992simple,sutton1999policy} & \tabincell{c}{Highway ramping;\\Figure-Eight} & \tabincell{l}{1. The typical Monte Carlo-based policy gradient algorithm. \\ 2. Stochastic gradient ascent is used to update model parameter. \\ 3. Large variance of gradient estimation. \\ 4. Poor model stability and low learning efficiency.}\\
\cmidrule{2-5}

~ & AC & \cite{konda1999actor} & \tabincell{c}{Highway ramping;\\Figure-Eight} &
\tabincell{l}{1. The DRL model consists of an actor network and a critic network. \\ \tabincell{l}{2. Actor network predicts the probability of the action, critic network predicts \\ \quad the value (Q value) in the current state.} \\ 3. Single-step update can be performed to achieve high update speed. \\ 4. Correlation exists between parameter updates of the two neural network. \\ \tabincell{l}{5. Difficult model convergence since the action generated by the actor network \\ \quad depends on the value predicted by the critic network}}\\
\cmidrule{2-5}

~ & A2C & \cite{mnih2016asynchronous} & \tabincell{c}{Highway ramping;\\Figure-Eight} & \tabincell{l}{1. Add a baseline to the calculation of Q values based on AC method. \\ 2. Reduce numerical variation in the actor
networks. \\ 3. Strong model stability.}\\
\cmidrule{2-5}

~ & NAF & \cite{gu2016continuous} &\tabincell{c}{Figure-Eight}& \tabincell{l}{\tabincell{l}{1. Normalized advantage function is implemented to extend DQN method to  \\ \quad continuous action space.} \\ \tabincell{l}{2. Q value is decoupled into value function and advantage function, and \\ \quad advantage function is calculated based on the Cholesky decomposition.}}\\
\cmidrule{2-5}

~ & Double NAF & \cite{van2016deep,gu2016continuous} & \tabincell{c}{Figure-Eight} & \tabincell{l}{\tabincell{l}{1. Combining Double DQN method with NAF method to reduce the \\ \quad overestimation of Q values.}}\\
\cmidrule{2-5}

~ & DDPG & \cite{lillicrap2015continuous} & \tabincell{c}{Figure-Eight} & \tabincell{l}{1. The actor-critic framework is introduced based on the DQN method. \\ 2. Target networks are created for both actor and critic networks. \\ 3. Experience replay is established to ensure high sample efficiency. \\4. Deterministic policy is not conducive to action exploration. \\5. Overestimation of Q value generated by the critic network.}
\\
\cmidrule{2-5}

~ & TD3 & \cite{fujimoto2018addressing} & \tabincell{c}{Figure-Eight} & \tabincell{l}{\tabincell{l}{1. Double Q learning is implemented based on the DDPG method to reduce \\ \quad the overestimation of Q value.} \\ 2. Delay actor network updates for more stable training for actor network. \\ \tabincell{l}{3. Add noise to the output action from target actor network to increase \\ \quad the stability.}}
\\
\cmidrule{2-5}

~ & PPO & \cite{schulman2017proximal} & \tabincell{c}{Highway ramping;\\Figure-Eight} &
\tabincell{l}{\tabincell{l}{1. Importance sampling is applied to change the actor-critic framework \\ \quad from on-policy to an off-policy.} \\ \tabincell{l}{2. Clipped surrogate objective is proposed to limit the amplitude of policy \\ \quad updates to avoid excessive strategy deviations.} \\ 3. Multiple training updates with one sample. \\ 4. Good versatile and samples complexity.}\\

\bottomrule
\end{tabular}
\end{center}
\end{table*}

\subsection{Results and Discussion}
\begin{figure*}[h]
  \centering
  \includegraphics[scale=0.58]{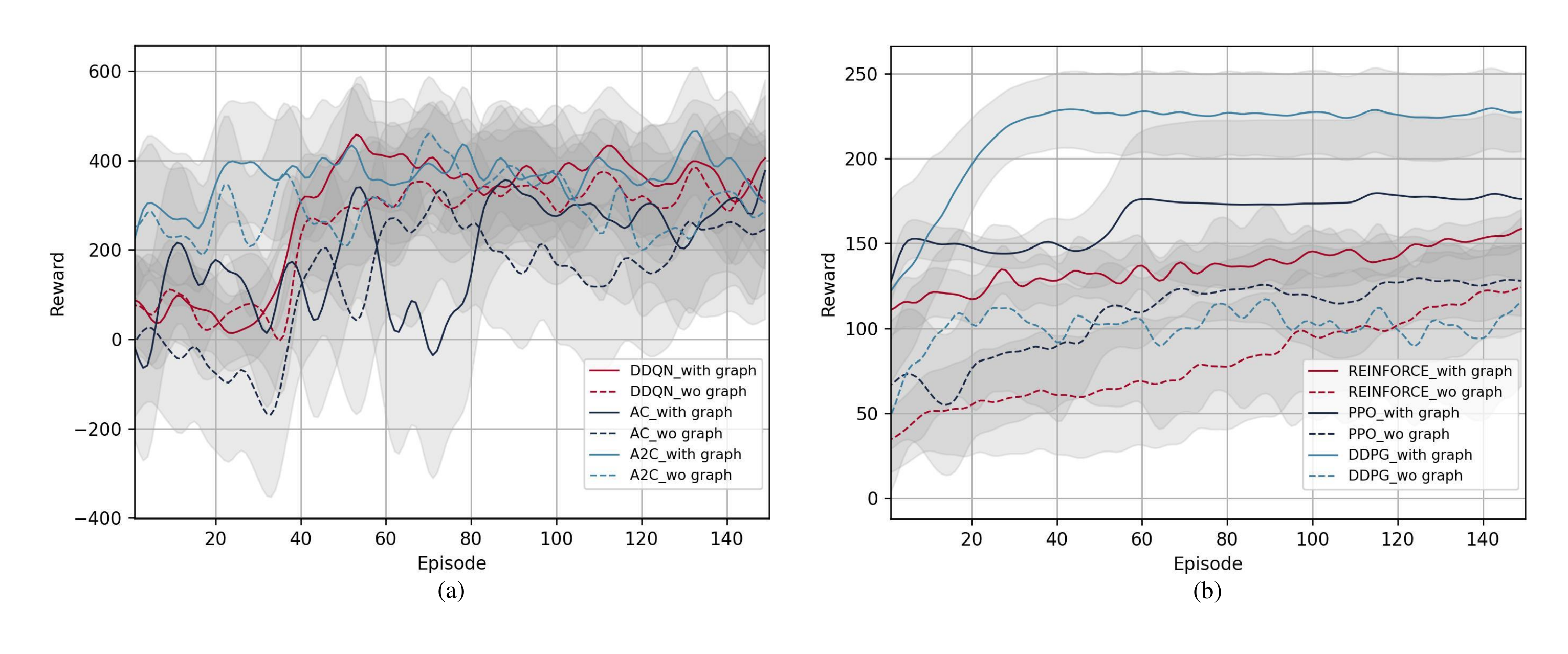}
  \caption{The curves of average training reward of several GRL-based methods and their corresponding DRL-based methods. The shaded areas show the standard deviation for three random seeds; the solid line represents the reward curve of a GRL-based method, while a dashed line represents the reward curve of a DRL-based method: (a) the reward curves of distributional DQN, AC, and A2C algorithm in the highway ramping scenario; (b) the reward curves of REINFORCE, PPO, and DDPG algorithm in the Figure-Eight scenario.}
  \label{figureReward}
\end{figure*}

The GRL methods implemented in the proposed framework are validated in their available scenarios. The ablation experiment was conducted to verify the effectiveness of the GRL-based methods, and their performances were compared in detail. Three random seeds were trained for 150 epochs, and the average training reward was calculated as the evaluation of a method. The performance of different GRL-based methods and their corresponding DRL-based methods are summarized in TABLE \ref{tab7}. The reward curves of several GRL-based methods in the two constructed scenarios are illustrated in Fig. \ref{figureReward}.

\begin{table*}[h]
\caption{Performance of Different GRL-based Methods.}
\label{tab7}
\begin{center}
\begin{tabular}{ccccc}
\toprule
Scenario & Method & Average reward of DRL & Average reward of GRL & Optimization rate (\%)\\

\midrule 
\multirow{17}{*}{Highway ramping} & DQN & 312.17 & 337.61 & 8.15 \\
\cmidrule{2-5}
~ & Double DQN & 337.79 & 374.51 & 10.87 \\
\cmidrule{2-5}
~ & Dueling DQN & 326.68 & 365.00 & 11.73 \\
\cmidrule{2-5}
~ & Noisy DQN & 325.21 & 367.03 & 12.86 \\
\cmidrule{2-5}
~ & DQN with PER & 307.98 & 349.64 & 13.53 \\
\cmidrule{2-5}
~ & Distributional DQN & 318.96 & 375.51 & 17.73 \\
\cmidrule{2-5}
~ & Rainbow DQN & 343.76 & 373.08 & 8.53 \\
\cmidrule{2-5}
~ & Reinforce & 268.13 & 321.30 & 19.83 \\
\cmidrule{2-5}
~ & AC & 132.40 & 192.63 & 45.49 \\
\cmidrule{2-5}
~ & A2C & 298.61 & 363.93 & 21.88 \\
\cmidrule{2-5}
~ & PPO & 324.75 & 376.64 & 15.98 \\

\midrule 

\multirow{12}{*}{Figure-Eight} & Reinforce & 79.95 & 135.81 & 69.87 \\
\cmidrule{2-5}
~ & AC & 142.42 & 187.26 & 31.48 \\
\cmidrule{2-5}
~ & A2C & 220.38 & 230.41 & 4.55 \\
\cmidrule{2-5}
~ & NAF & 59.32 & 62.72 & 5.74 \\
\cmidrule{2-5}
~ & Double NAF & 59.25 & 62.70 & 5.82 \\
\cmidrule{2-5}
~ & DDPG & 100.77 & 215.56 & 113.92 \\
\cmidrule{2-5}
~ & TD3 & 123.25 & 205.31 & 66.58 \\
\cmidrule{2-5}
~ & PPO & 107.35 & 165.12 & 53.81 \\

\bottomrule
\end{tabular}
\end{center}
\end{table*}

\subsubsection{Ablation experiment}
The experimental results  showed that all GRL-based methods had higher average rewards than their corresponding DRL-based methods in both test scenarios. This indicated that implementing graphic techniques into the DRL methods could improve the multi-agent decision-making performance of CAVs in mixed autonomy traffic.

It should be noted that different DRL-based methods have different effects on optimization. After the advent of graphic technology, the optimization effect of policy-based has generally outperformed the value-based methods. In the highway ramping scenario, the optimization effect of the DQN series methods was not as apparent as that of the policy-based methods. In the Figure-Eight  scenario, most algorithms achieved good optimization results. However, the optimization rate of the A2C was negligible because the DRL framework already achieved a better average training reward in this scenario compared to the other methods. Moreover, it is crucial to note that the NAF and DoubleNAF algorithms had almost no optimization effect. Considering that these two algorithms denote improved versions of the classical value-based approach of DQN, it could be further concluded that the optimization effect of the value-based methods was not noticeable.

In the comparative study, different types of traffic scenarios showed different overall optimization effects. The overall optimization effect of the Figure-Eight scenario was better than that of the highway ramping scenario. This could be because, in the Figure-Eight scenario, the mutual effect of vehicles was more obvious. Namely, unreasonable driving behavior of any vehicle affected the speed of other vehicles, thus impacting the overall traffic safety and efficiency. However, in the highway ramping scenario, the distribution of vehicles was relatively sparse, and the driving behavior of vehicles did not significantly affect the regular operation of the other vehicles. Thus, it can be concluded that the GRL-based methods are more suitable for handling decision-making problems in a traffic scenario where collaboration between vehicles is required to ensure driving safety. 

\subsubsection{Comparison of different GRL methods}
The experimental results showed that different methods had different performances in the two test scenarios. In the highway ramping scenario, several improved DQN methods achieved different degrees of optimization effect than the original DQN algorithm in both DRL- and GRL-based frameworks. The A2C and PPO algorithms performed well in both frameworks; the optimization effect of the GRL-based methods compared to the DRL-based methods was apparent. However, the AC algorithm performed poorly. From the holistic perspective, the value-based methods performed better than the policy-based methods in both frameworks.

In the Figure-Eight scenario, the NAF and DoubleNAF algorithms performed well in neither the DRL- not GRL-based frameworks, indicating that the extension of the classical DQN algorithm to the continuous action space could not achieve the expected results. The A2C performed well in this scenario, and both the DRL- and GRL-based frameworks achieved high reward values. The performance of other algorithms didn't differ much in the DRL-based framework. Still, their performances differed significantly from those in the GRL-based framework, and the optimization effect was significant.


\section{Challenges and Future Outlook}
This section identifies the current challenges and presents future research directions in the field of GRL-based methods for decision-making in mixed autonomy traffic based on the results of the state-of-the-art research and comparative study presented in this paper.

\subsection{Communication}
The data sharing among CAVs at the perceptual information level and their collaboration at the motion planning level rely on efficient communication between vehicles and data transmission between vehicles and centralized controllers. In \cite{liu2022integrated}, the authors discussed the influence of packet loss on the information transmission in distributed vehicle control problems. In \cite{zhou2021multiagent}, the impact on collaborative multi-agent decision-making in communication failure scenarios was investigated. With the development of 5G technology, low-power and low-latency technologies have significantly developed. Further, in \cite{wang2022software}, it was investigated how to model communication mechanisms in intelligent transportation systems using GNNs. The model constructed in this paper does not consider the effects of information loss, errors, and delays in vehicle communication on the results. In \cite{blumenkamp2022framework}, the authors modeled the effect of information communication and delay on real-world multi-robot systems using the graph structure. The structured modeling with a GNN facilitates the simulation and processing under more realistic future traffic conditions. A future direction regarding the proposed GRL-based framework could be to model the effect of communication between vehicles in the collaborative decision-making model.

\subsection{Reward Design}
In the introduced  decision-making model based on DRL, the final model performance depends on the reward function design in the DRL module and the weight distribution of the rewards \cite{ha2020leveraging,chen2021graph}. Therefore, an implementation strategy of the requirements for model performance into the designed reward functions can significantly affect the training results \cite{chen2020deep}. For instance, in the considered highway ramping and Figure-Eight scenarios, the definition of reward was influenced by the scenario and task performances (e.g., the overall traffic efficiency, traffic efficiency in a particular lane, reduction in the passage time of a specific type of vehicle in the scenario). Moreover, for the cooperative multi-agent decision-making problem in the mixed autonomy traffic, the conflict between the overall reward and the individual reward must be considered. This includes social interaction and implicit synergy between human drivers with different levels of aggressiveness \cite{ha2020leveraging}. The design process of the reward function also needs to consider the priority of HVs and AVs, and such priorities need to be taken into account in the design of the loss function, and the development and robustness of laws and regulations involving autonomous driving.

\subsection{Transfer Learning}
The reviewed approaches and the proposed framework focus on the multi-agent collaborative approach based on RL. According to \cite{taylor2009transfer}, when the distribution of scenarios where models are trained and test scenarios have certain differences, the performance of the RL-based models decreases. To solve the migration learning problem involved in model adaptation to unseen traffic scenarios, the constructed models need to be adapted to the new scenarios encountered in tests. However, this can degrade model performance in historical scenarios learned during the training process, which is referred to as catastrophic forgetting. Therefore, one of the current work challenges is making the models capable of continuous learning and evolution. Moreover, the designed models need to be stable, incremental, and efficient in adapting to new scenarios and environments following the changes in a model’s test and actual traffic scenarios \cite{lesort2020continual}.

\subsection{Human Factor}
According to \cite{he2022modelling}, human drivers and passengers in AVs have different levels of risk acceptance and perception. Current research has not adequately considered the human perceived risk for driver and passenger comfort. In \cite{kolekar2020human}, it was pointed out that the risk range around a vehicle should be a risk field related to the driver’s risk acceptance level. Therefore, the developed model needs to consider the individualized risk acceptance levels. Meanwhile, in complex and intense interaction environments and scenarios, the interactions between human drivers and between self-driving vehicles and HVs are different, so the variability brought by the human factor should be considered.

\subsection{Traffic Control System Cooperative Feature}
In this study, traffic signals, which are widely available in urban road networks, have not been modeled in the current phase of the model. However, in future intelligent transportation systems, the optimization of vehicle decision control behavior and traffic signal control phases should be a coupled model. In the case of complete vehicle-road information, CAVs on urban roads can accurately sense and predict the signal timing scheme of downstream intersections and adjust the trajectory accordingly. The signals can also actively provide real-time information on the position and speed of arriving vehicles and optimize the signal parameters. Thus, fleet trajectory control and traffic signal optimization are interdependent and affect each other. Namely, the trajectory control or signal optimization alone cannot significantly improve the intersection capacity and traffic flow operation efficiency. Only collaborative optimization of intelligent, networked fleet trajectory and traffic signal can achieve the goals of minimum delay, stop times, fuel consumption, emissions, and optimal traffic efficiency \cite{liu2022single}. Therefore, future research should focus on the three following points: (1) how to design the trajectory control algorithm and strategy for the intelligent networked fleet so that vehicles can slow down smoothly in the face of red-light signals to achieve the minimum number of stops, fuel consumption, and emissions; (2) how to make full use of fleet information to optimize the signal timing scheme to achieve the control objectives of minimum delay and optimal traffic efficiency; (3) how to be compatible with upstream and downstream intersections to extend the optimization control to the road network and solve the optimization problem in real-time.

\subsection{Uncertainty Problem}
The performance of machine learning-based models greatly depends on the training data selection, which can result in accidental and cognitive uncertainties. The accidental uncertainty (non-reducible) is caused by the inherent randomness in the collected data, which includes the imperfection of information sharing in mixed autonomy traffic and sensor shift \cite{gawlikowski2021survey}. For instance, in a lane-changing scenario, AVs may not be able to obtain the turn signal of the surrounding HVs accurately, which will result in a decrease in the passable area, which in turn affects the decision-making process of AVs in a traffic environment. Cognitive uncertainty, which is also known as knowledge uncertainty, originates from the rare cases in the test environment, which denote cases that rarely occur in the training environment; this problem can be addressed by increasing the amount of training data. The proposed model does not systematically consider the effects of the two uncertainties on the model performance. Also, in this work, it has not been analyzed how to reduce the cognitive uncertainty and adapt to the new test environment through the model structure adjustment. In future work, a multi-agent GRL-based model for the mixed autonomy traffic considering uncertainty could be considered.

\subsection{Coordination of Global and Local Information}
The model constructed in this study assumes that there is communication between all AVs and that the sensing ability of a sensor is within a defined perception range. In the graph structure design, communication between vehicles and sensing range can be reflected by the connectivity between nodes in a graph. In future work, the impact of the perception range and communication range of AVs on the overall performance of the model could be analyzed based on a GRL-based framework. In addition, it could be discussed whether the requirement for the sensing range of a single AV could be decreased to reduce the sensor cost and avoid the uncertainty problem in remote sensing.

\subsection{Vehicle Models}
The proposed GRL model for decision-making in this paper uses a simplified vehicle kinematics model configured by \cite{7995802}. Accordingly, the action space of RL considers only several discrete decision behaviors, including lane-changing commands. However, in this work, the whole lane-changing process is represented by an ideal model. Still, more complex vehicle kinematics and dynamics models should be  considered in the future, since the road conditions and the parameters of a vehicle model are crucial for an accurate evaluation of vehicle motion.

\section{Conclusion}
This paper studies the GRL-based methods for multi-agent decision-making in mixed autonomy traffic. A generic and modular GRL-based framework is designed, and the techniques of each module in the proposed framework are elaborated. A review of state-of-the-art DRL- and GRL-based approaches for solving decision-making problems in mixed autonomy traffic is provided, focusing on the recent research topics. A comparative study is conducted based on the GRL-based framework, and an open-source code is provided. Finally, the current challenges and future research directions in the field of GRL-based methods are summarized. This  work has great potential in the design of GRL-based methods for multi-agent decision-making in mixed autonomy traffic and provides a research basis for related researchers. Future work will focus on constructing a more comprehensive GRL framework according to the proposed research challenges.

\section{Acknowledgment}
The authors would like to thank Yuxi Liang for the help with the development of the program library, and we also thank Tian Luan for the collation of experimental results in the comparative study part of this paper.

\small
\bibliographystyle{IEEEtran}
\bibliography{myref}

\vfill

\end{document}